\documentclass[letterpaper, 10 pt, conference]{ieeeconf}  

\IEEEoverridecommandlockouts                              

\usepackage{cite}
\usepackage[german]{babel}

\usepackage{graphicx} 
\usepackage{amsmath} 
\usepackage{amssymb}  
\usepackage{algorithm,algorithmic}
\usepackage{hyperref}
\usepackage{tikz}
\usepackage[caption=false]{subfig}

\usetikzlibrary{positioning, shapes.geometric, arrows.meta,arrows}

\title{\LARGE \bf
Let Hybrid \text{A}$^{\negthinspace*}$  Path Planner Obey Traffic Rules: \\A Deep Reinforcement Learning-Based Planning Framework 
}

\author{Xibo Li$^{*}$, Shruti Patel$^{*\dagger}$ and Christof B\"uskens$^{*\dagger}$
	\thanks{$^{*}$The authors are with the working group Optimization and Optimal Control, Center for Industrial Mathematics, University of Bremen, 28359 Bremen, Germany, contact {\tt\small lixibo@uni-bremen.de}  }%
	\thanks{$^{\dagger}$The authors are with TOPAS Industriemathematik Innovation gGmbH, 28359 Bremen, Germany, contact {\tt\small shruti.patel@topas.tech}  }%
	\thanks{\textit{Corresponding author: Xibo Li}}%
}%

\begin{document}

\maketitle
\thispagestyle{empty}
\pagestyle{empty}

\begin{abstract}
	
Deep reinforcement learning (DRL) allows a system to interact with its environment and take actions by training an efficient policy that maximizes self-defined rewards. In autonomous driving, it can be used as a strategy for high-level decision making, whereas low-level algorithms such as the hybrid \text{A}$^{\negthinspace*}$ path planning have proven their ability to solve the local trajectory planning problem. In this work, we combine these two methods where the DRL makes high-level decisions such as lane change commands. After obtaining the lane change command, the hybrid \text{A}$^{\negthinspace*}$ planner is able to generate a collision-free trajectory to be executed by a model predictive controller (MPC). In addition, the DRL algorithm is able to keep the lane change command consistent within a chosen time-period. Traffic rules are implemented using linear temporal logic (LTL), which is then utilized as a reward function in DRL. Furthermore, we validate the proposed method on a real system to demonstrate its feasibility from simulation to implementation on real hardware.

\end{abstract}


\section{INTRODUCTION}
\label{sec:introduction}

Over the past few years, Advanced Driver Assistance Systems (ADAS) have improved road transportation safety and efficiency \cite{jimenez2016advanced}. 
Autonomous driving, considered to be the next step of ADAS, has also made great progress due to the development of sensor technology, control theory, and deep learning \cite{muhammad2020deep}.
Despite the advances in autonomous driving, when and how to choose the current driving behavior is still quite a challenging task in complex driving scenarios \cite{badue2021self}. 
Rule-based methods are usually implemented in order to solve straightforward scenarios such as lane change, lane merging, emergency braking, and so on. 
However, these hand-crafted rules can be inefficient in dense traffic with multiple moving vehicles. 

Recently, reinforcement learning (RL) has shown its potential in solving complicated tasks and achieving a better performance than human experts while playing Atari games \cite{mnih2013playing} and the game of Go \cite{silver2017mastering}. 
Meanwhile, those RL methods are also extended in behavior planning in order to make feasible decisions in different driving scenarios \cite{hoel2018automated}\cite{wang2018reinforcement}. 
Simulation tools such as SUMO \cite{krajzewicz2012recent} and CARLA \cite{dosovitskiy2017carla} are used to construct various driving scenarios in order to train the RL agent. Referring to real-world implementations, mainly end-to-end approaches have been applied \cite{kendall2019learning}\cite{folkers2019controlling}. However, several challenges persist when applying RL methods to autonomous driving.

\begin{figure}[t]
	\centering
	\includegraphics[width=1\linewidth]{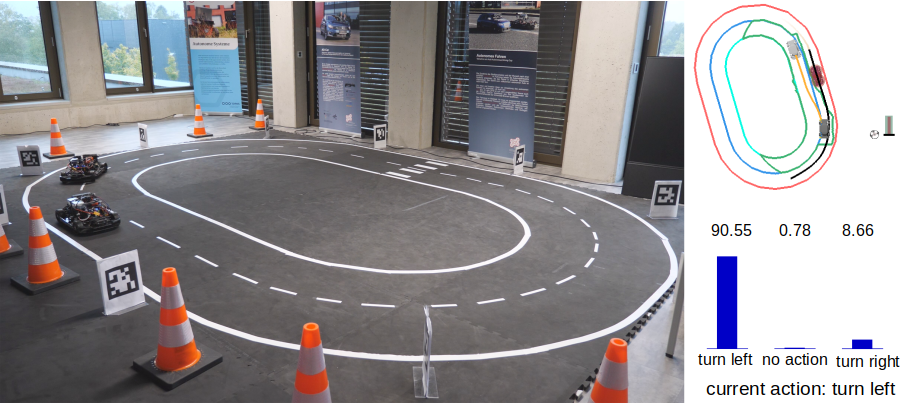}
	\caption{Clockwise from the left: two ADAS model cars driving cooperatively, with the high-level decision being made by RL; real-time trajectory planning in the visualization; and the action distribution of the RL Agent after training.}
	\label{fig:overview}
\end{figure}

Firstly, the RL policy trained in simulation is hard to directly apply in real-world situations because the simulation environment does not adequately mirror real-world conditions. Secondly, when the steering and acceleration are generated by end-to-end approaches, a reliable, smooth, and comfortable trajectory is hard to ensure. Thirdly, the RL agent is not able to obey basic traffic rules. To address the first challenge, we propose an RL platform based on the EB Assist Automotive Data and Time-Triggered Framework (ADTF), commonly used in the automotive industry \cite{hellmund2016robot}. The RL agent is initially trained in the ADTF simulation environment and then directly implemented on an ADAS model car, as depicted in Fig. \ref{fig:overview}. Further  details of the model car can be found in \cite{li2023indoor}. To tackle the second challenge, we propose a framework for combining deep reinforcement learning with local trajectory planning. The proposed framework is structured into two components, \emph{high-level behavior planning} and \emph{low-level trajectory planning}. \emph{High-level behavior planning} employs DRL methods to generate lane change commands, while the hybrid \text{A}$^{\negthinspace*}$ planner serves as the \emph{low-level trajectory planner} tasked with generating the local trajectory \cite{folkers2020time}. Specifically, the Proximal Policy Optimization (PPO) \cite{schulman2017proximal} algorithm is implemented to generate discrete lane change actions, aiding the local trajectory planning module in generating reliable and efficient lane change trajectories.  As for the third challenge, the RL agent is trained to comply with traffic rules, which are encoded using LTL and incorporated into the agent's reward function \cite{esterle2020formalizing}\cite{wang2022learning}. This approach ensures that our agent obeys traffic rules when making decisions.

Concretely, the main contributions of our work are summarized as follows: 
\begin{enumerate}
	\item We propose a novel hierarchical decision-making framework named Time-Period Reinforcement Learning (TPRL). With TPRL, the action can be sampled within fixed time period and remain consistent for the duration of each period. As a hierarchical framework, it combines the Hybrid \text{A}$^{\negthinspace*}$ trajectory planning algorithm with a PPO-based deep reinforcement learning approach and ensures a collision-free and executable trajectory in a lane change scenario.
	\item  We encode the LTL into the reward function, which
	enables the RL agent to learn and adhere to traffic
	rules and make reasonable decisions.
	\item The gap between simulation and real-world implementation is minimized by directly training the RL algorithm on ADTF, where the proposed method can be easily validated in realistic scenarios. A short video demonstrating real-world cooperative driving with our TPRL algorithm is available at \url{https://youtu.be/S06FYFLqvC0}. 
\end{enumerate}

\section{RELATED WORKS }
\label{sec:related_works}

\subsection{Heuristic-based Behavior Planning}

Heuristic-based behavior planning methods attempt to solve the behavior planning problem by developing a set of rules for various driving scenarios. For example, in \cite{urmson2008autonomous}, a rule-based behavioral layer with three sub-components: lane driving, intersection handling, and goal selection, is designed in order to make tactical decisions. 
Additionally, other heuristic methods such as decision trees \cite{claussmann2015path} or finite state machines \cite{montemerlo2008junior} are also explored in other autonomous driving systems.
However, these methods are primarily employed to tackle limited, uncomplicated scenarios. As the complexity of driving situations increases, the effectiveness of hand-crafted rules diminishes due to the challenges associated with managing their growing complexity and tuning hyper-parameters.

\subsection{Learning-based Behavior Planning}

By interacting with the environment, learning-based methods enable a system to execute actions based on pre-defined rewards.
Deep Q Network (DQN) is first implemented to execute reasonable lane change decisions across various simulated environments  \cite{hoel2018automated} \cite{wang2018reinforcement}. 
In addition to DQN, safe, efficient, and comfortable lane change decisions are executed using PPO in mandatory lane change scenarios within SUMO Simulation \cite{ye2020automated}, where in \cite{krasowski2020safe} safe lane change decisions are ensured by designing an additional safety layer between the agent and the environment. 
By combining RL with Imitation Learning, \cite{zhang2021end} employs bird's eye view images as the state input and achieves impressive performance for different traffic scenarios such as the stop signs, lane changing, merging, and US-style junctions in CARLA Simulation. 
Another imitation learning method, Adversarial Inverse Reinforcement Learning (AIRL) is utilized in \cite{wang2021decision} in order to learn both a behavioral policy and a reward function. 
For the autonomous overtaking problem in high-speed situations, \cite{song2021autonomous} implements curriculum reinforcement learning in a racing game and achieves a comparable performance to the experienced human driver. Besides the end-to-end methods, the authors of \cite{qiao2021behavior}  \cite{li2021safe} also implement some hierarchical structures in simulated urban environments. 
These previous works have shown the potential of using learning-based methods to solve the behavior planning problem. 
However, very few combine learning-based methods together with local trajectory planning. Most of the works only test the performance in simulation environments. 
In this paper, we present a comprehensive learning-based framework  that combines PPO with trajectory planning and control.
Furthermore, the performance is also validated both in simulation and real-world situations.

\section{PROBLEM DESCRIPTION }
\label{sec:problem_description}

A lane change action in road traffic can occur either mandatorily or discretely. 
In the mandatory scenario, the decision is typically made when exiting the highway or merging from multiple lanes into a single lane. 
These situations involve a high volume of vehicles with a complex traffic flow. 
To simplify the scenario, we initially demonstrate it with two vehicles before expanding to more complex situations.
Therefore, in this paper, we primarily focus on the discretionary lane change scenario, which is suitable for testing with two ADAS model cars.

Here we formulate the discretionary lane change scenario as a partially observed Markov decision process \cite{spaan2012partially} defined by a tuple $< \mathcal S, \mathcal A, \mathcal O, \mathcal T,  \mathcal E,  \mathcal R, \gamma >$. The state, action and observation spaces are represented by $ \mathcal S $ , $\mathcal A$ and $\mathcal O$ , where the state vector at timestep $t$ is denoted as $ s_t \in \mathcal S $, the action taken by the ego agent at timestep $t$ is defined as $ a_t \in \mathcal A $ and the observation received by the ego agent at timestep $t$ is $ o_t \in \mathcal O $. The transition model  $ \mathcal T $  at timestep $t$ is defined as $ p_t(s_{t+1}|s_t, a_t) $, where $p_t$ denotes the current transition probability from state-action pair  $ (s_t, a_t) $ to the next state $s_{t+1}$.  The emission probability $ \mathcal E $ at timestep $t$ is defined as  $ p_t(o_{t}|s_t) $, which represents the probability of an observation $o_{t}$ being generated from a state $s_{t}$. The reward function is defined by $ r(s_t, a_t) \in \mathcal R $ and the discount factor is denoted by  $ \gamma \in [0,1] $. The overall objective function of reinforcement learning is to maximize the expected discounted return by finding an optimal policy $\pi^*$:
\begin{align}
	\pi^* = \arg \max_{\pi}  \mathop{{}\mathbb{E}} \left[  \sum_{t=0}^{\infty} \gamma^t r(s_t, a_t)  \right],
	\label{eq:policy_objective_function}
\end{align}

\section{METHODOLOGY}
\label{sec:method}

\subsection{Overview of the Reinforecement Learning Framework in ADTF}
\begin{figure}[t]
	\centering
	
	\tikzset{every picture/.style={line width=0.75pt}} 
	
	\begin{tikzpicture}[x=0.5pt,y=0.5pt,yscale=-1,xscale=1]
	
	\draw (240.41,307.11) node  {\includegraphics[width=52.37pt,height=33.58pt]{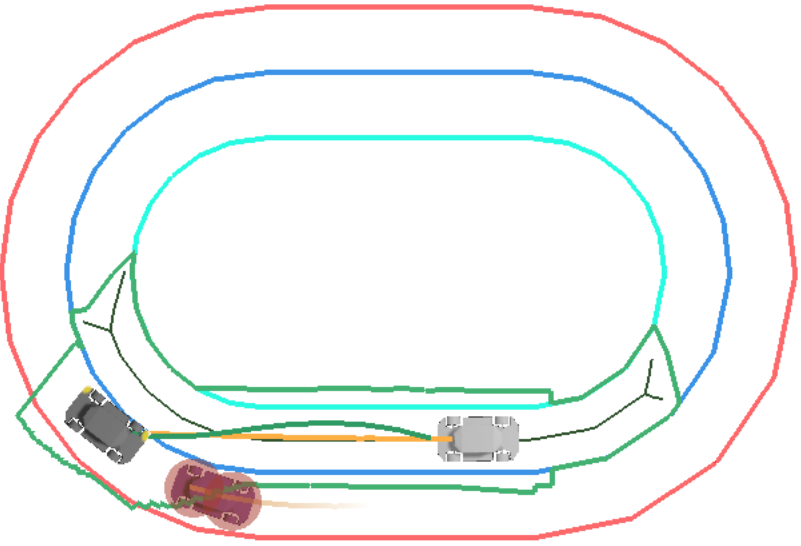}};
	\draw   (169.4,247.8) -- (320.2,247.8) -- (320.2,346.57) -- (169.4,346.57) -- cycle ;
	\draw   (321,299) -- (351,299) ;
	\draw   [shift={(351,299)}, rotate = 180] [fill={rgb, 255:red, 0; green, 0; blue, 0 }  ][line width=0.08]  [draw opacity=0] (10.72,-5.15) -- (0,0) -- (10.72,5.15) -- (7.12,0) -- cycle    ;
	\draw  [fill={rgb, 255:red, 255; green, 255; blue, 255 }  ,fill opacity=1 ] (350.8,257.1) -- (470.8,257.1) -- (470.8,327) -- (350.8,327) -- cycle ;
	\draw    (356,271.41) -- (465.75,271.6) ;
	\draw   (355.67,310.78) -- (464.73,311.06) ;
	\draw  [dash pattern={on 4.5pt off 4.5pt}]  (356.69,289.32) -- (465.75,289.6) ;
	\draw [color={rgb, 255:red, 74; green, 144; blue, 226 }  ,draw opacity=1 ][line width=1.5]    (383.52,301.43) .. controls (430.4,300.96) and (431.56,280.33) .. (466.03,280.33) ;
	\draw (369.79,300.08) node [rotate=-90.95,xslant=-0.01] {\includegraphics[width=7.19pt,height=14.24pt]{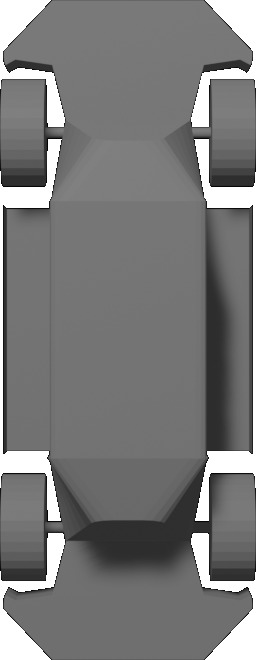}};
	\draw (449.59,300.33) node [rotate=-89.31,xslant=0.01] {\includegraphics[width=7.19pt,height=14.24pt]{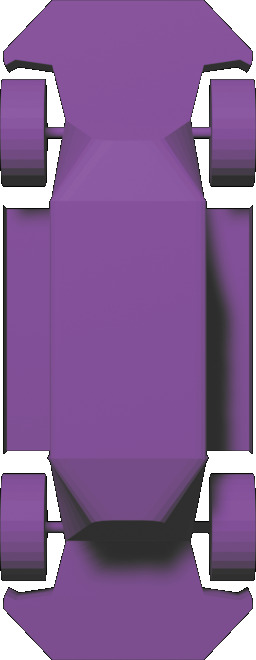}};
	
	\draw  [fill={rgb, 255:red, 255; green, 255; blue, 255 }  ,fill opacity=1 ] (360.8,264.1) -- (480.8,264.1) -- (480.8,334) -- (360.8,334) -- cycle ;
	\draw    (366,278.41) -- (475.75,278.6) ;
	\draw    (365.67,317.78) -- (474.73,318.06) ;
	\draw  [dash pattern={on 4.5pt off 4.5pt}]  (366.69,296.32) -- (475.75,296.6) ;
	\draw [color={rgb, 255:red, 74; green, 144; blue, 226 }  ,draw opacity=1 ][line width=1.5]    (393.52,308.43) .. controls (440.4,307.96) and (441.56,287.33) .. (476.03,287.33) ;
	\draw (394.79,307.08) node [rotate=-90.95,xslant=-0.01] {\includegraphics[width=7.19pt,height=14.24pt]{figures/car.jpg}};
	\draw (459.59,307.33) node [rotate=-89.31,xslant=0.01] {\includegraphics[width=7.19pt,height=14.24pt]{figures/moving_object_car.jpg}};
	
	\draw  [fill={rgb, 255:red, 255; green, 255; blue, 255 }  ,fill opacity=1 ] (371.47,273) -- (500,273) -- (500,348) -- (371.47,348) -- cycle ;
	\draw (423.21,317.45) node [rotate=-78.15,xslant=-0.03] {\includegraphics[width=7.19pt,height=14.24pt]{figures/car.jpg}};
	\draw [color={rgb, 255:red, 74; green, 144; blue, 226 }  ,draw opacity=1 ][line width=1.5]    (436.77,314.1) .. controls (442.7,314.8) and (465.11,297.41) .. (488.83,297.41) ;
	\draw  (382.37,287.93) -- (490.85,288.13) ;
	\draw    (382.05,329) -- (489.85,329.29) ;
	\draw  [dash pattern={on 4.5pt off 4.5pt}]  (383.05,306.61) -- (490.85,306.9) ;
	\draw (474.88,318.1) node [rotate=-89.35,xslant=0.01] {\includegraphics[width=7.19pt,height=14.24pt]{figures/moving_object_car.jpg}};
	
	\draw   (409.8,254) -- (409.8,212) ;
	\draw [shift={(409.8,208)}, rotate = 90] [fill={rgb, 255:red,0; green, 0; blue, 0 }  ][line width=0.08]  [draw opacity=0] (10.72,-5.15) -- (0,0) -- (10.72,5.15) -- (7.12,0) -- cycle    ;
	\draw    (409.8,351) -- (409.8,385) ;
	\draw [shift={(409.8,388)}, rotate = 270] [fill={rgb, 255:red, 0; green, 0; blue, 0 }  ][line width=0.08]  [draw opacity=0] (10.72,-5.15) -- (0,0) -- (10.72,5.15) -- (7.12,0) -- cycle    ;
	\draw  [color={rgb, 255:red, 74; green, 144; blue, 226 }  ,draw opacity=1 ][fill={rgb, 255:red, 74; green, 144; blue, 226 }  ,fill opacity=1 ] (357.95,136.88) .. controls (357.95,133.02) and (361.43,129.89) .. (365.72,129.89) .. controls (370.02,129.89) and (373.5,133.02) .. (373.5,136.88) .. controls (373.5,140.75) and (370.02,143.88) .. (365.72,143.88) .. controls (361.43,143.88) and (357.95,140.75) .. (357.95,136.88) -- cycle ;
	\draw  [color={rgb, 255:red, 74; green, 144; blue, 226 }  ,draw opacity=1 ][fill={rgb, 255:red, 74; green, 144; blue, 226 }  ,fill opacity=1 ] (357.68,157.54) .. controls (357.68,153.67) and (361.16,150.54) .. (365.46,150.54) .. controls (369.75,150.54) and (373.24,153.67) .. (373.24,157.54) .. controls (373.24,161.4) and (369.75,164.53) .. (365.46,164.53) .. controls (361.16,164.53) and (357.68,161.4) .. (357.68,157.54) -- cycle ;
	\draw  [color={rgb, 255:red, 74; green, 144; blue, 226 }  ,draw opacity=1 ][fill={rgb, 255:red, 74; green, 144; blue, 226 }  ,fill opacity=1 ] (357.95,178.9) .. controls (357.95,175.04) and (361.43,171.91) .. (365.72,171.91) .. controls (370.02,171.91) and (373.5,175.04) .. (373.5,178.9) .. controls (373.5,182.76) and (370.02,185.89) .. (365.72,185.89) .. controls (361.43,185.89) and (357.95,182.76) .. (357.95,178.9) -- cycle ;
	\draw  [color={rgb, 255:red, 74; green, 144; blue, 226 }  ,draw opacity=1 ][fill={rgb, 255:red, 74; green, 144; blue, 226 }  ,fill opacity=1 ] (389.36,125.02) .. controls (389.36,121.15) and (392.84,118.02) .. (397.14,118.02) .. controls (401.43,118.02) and (404.91,121.15) .. (404.91,125.02) .. controls (404.91,128.88) and (401.43,132.01) .. (397.14,132.01) .. controls (392.84,132.01) and (389.36,128.88) .. (389.36,125.02) -- cycle ;
	\draw  [color={rgb, 255:red, 74; green, 144; blue, 226 }  ,draw opacity=1 ][fill={rgb, 255:red, 74; green, 144; blue, 226 }  ,fill opacity=1 ] (389.36,146.38) .. controls (389.36,142.52) and (392.84,139.39) .. (397.14,139.39) .. controls (401.43,139.39) and (404.91,142.52) .. (404.91,146.38) .. controls (404.91,150.24) and (401.43,153.37) .. (397.14,153.37) .. controls (392.84,153.37) and (389.36,150.24) .. (389.36,146.38) -- cycle ;
	\draw  [color={rgb, 255:red, 74; green, 144; blue, 226 }  ,draw opacity=1 ][fill={rgb, 255:red, 74; green, 144; blue, 226 }  ,fill opacity=1 ] (389.62,167.98) .. controls (389.62,164.12) and (393.11,160.99) .. (397.4,160.99) .. controls (401.7,160.99) and (405.18,164.12) .. (405.18,167.98) .. controls (405.18,171.84) and (401.7,174.97) .. (397.4,174.97) .. controls (393.11,174.97) and (389.62,171.84) .. (389.62,167.98) -- cycle ;
	\draw  [color={rgb, 255:red, 74; green, 144; blue, 226 }  ,draw opacity=1 ][fill={rgb, 255:red, 74; green, 144; blue, 226 }  ,fill opacity=1 ] (389.89,189.11) .. controls (389.89,185.25) and (393.37,182.11) .. (397.66,182.11) .. controls (401.96,182.11) and (405.44,185.25) .. (405.44,189.11) .. controls (405.44,192.97) and (401.96,196.1) .. (397.66,196.1) .. controls (393.37,196.1) and (389.89,192.97) .. (389.89,189.11) -- cycle ;
	\draw  [color={rgb, 255:red, 74; green, 144; blue, 226 }  ,draw opacity=1 ][fill={rgb, 255:red, 74; green, 144; blue, 226 }  ,fill opacity=1 ] (421.57,137.36) .. controls (421.57,133.5) and (425.05,130.37) .. (429.34,130.37) .. controls (433.64,130.37) and (437.12,133.5) .. (437.12,137.36) .. controls (437.12,141.22) and (433.64,144.35) .. (429.34,144.35) .. controls (425.05,144.35) and (421.57,141.22) .. (421.57,137.36) -- cycle ;
	\draw  [color={rgb, 255:red, 74; green, 144; blue, 226 }  ,draw opacity=1 ][fill={rgb, 255:red, 74; green, 144; blue, 226 }  ,fill opacity=1 ] (421.3,158.72) .. controls (421.3,154.86) and (424.78,151.73) .. (429.08,151.73) .. controls (433.37,151.73) and (436.86,154.86) .. (436.86,158.72) .. controls (436.86,162.59) and (433.37,165.72) .. (429.08,165.72) .. controls (424.78,165.72) and (421.3,162.59) .. (421.3,158.72) -- cycle ;
	\draw  [color={rgb, 255:red, 74; green, 144; blue, 226 }  ,draw opacity=1 ][fill={rgb, 255:red, 74; green, 144; blue, 226 }  ,fill opacity=1 ] (421.57,180.09) .. controls (421.57,176.22) and (425.05,173.09) .. (429.34,173.09) .. controls (433.64,173.09) and (437.12,176.22) .. (437.12,180.09) .. controls (437.12,183.95) and (433.64,187.08) .. (429.34,187.08) .. controls (425.05,187.08) and (421.57,183.95) .. (421.57,180.09) -- cycle ;
	\draw  [color={rgb, 255:red, 74; green, 144; blue, 226 }  ,draw opacity=1 ][fill={rgb, 255:red, 74; green, 144; blue, 226 }  ,fill opacity=1 ] (452.98,158.01) .. controls (452.98,154.15) and (456.46,151.02) .. (460.76,151.02) .. controls (465.05,151.02) and (468.53,154.15) .. (468.53,158.01) .. controls (468.53,161.87) and (465.05,165) .. (460.76,165) .. controls (456.46,165) and (452.98,161.87) .. (452.98,158.01) -- cycle ;
	\draw    (373.5,136.88) -- (389.36,125.02) ;
	\draw    (373.5,136.88) -- (389.36,146.38) ;
	\draw    (373.5,136.88) -- (389.62,167.98) ;
	\draw    (373.5,136.88) -- (389.47,188.29) ;
	\draw    (373.24,157.54) -- (389.36,125.02) ;
	\draw    (373.24,157.54) -- (389.36,146.38) ;
	\draw    (373.24,157.54) -- (389.62,167.98) ;
	\draw    (373.24,157.54) -- (389.89,189.11) ;
	\draw    (373.5,178.9) -- (389.36,125.02) ;
	\draw    (373.5,178.9) -- (389.36,146.38) ;
	\draw    (373.5,178.9) -- (389.62,167.98) ;
	\draw    (373.5,178.9) -- (389.89,189.11) ;
	\draw    (404.91,125.02) -- (421.99,138.01) ;
	\draw    (404.91,146.38) -- (421.99,138.01) ;
	\draw    (405.18,167.98) -- (421.65,138.87) ;
	\draw    (405.44,189.11) -- (421.99,138.01) ;
	\draw    (404.91,125.02) -- (421.3,158.01) ;
	\draw    (404.91,146.38) -- (421.3,158.01) ;
	\draw    (405.18,167.98) -- (421.3,158.01) ;
	\draw    (405.44,189.11) -- (421.3,158.01) ;
	\draw    (421.57,179.37) -- (404.91,125.02) ;
	\draw    (404.91,146.38) -- (421.57,179.37) ;
	\draw    (405.18,167.98) -- (421.57,179.37) ;
	\draw    (405.44,189.11) -- (421.57,179.37) ;
	\draw    (437.12,136.65) -- (452.98,158.01) ;
	\draw    (436.86,158.01) -- (452.98,158.01) ;
	\draw    (437.12,179.37) -- (452.98,158.01) ;
	\draw    (336.93,110) -- (493,110) -- (493,204) -- (336.93,204) -- cycle ;
	\draw  [color={rgb, 255:red, 245; green, 166; blue, 35 }  ,draw opacity=1 ][fill={rgb, 255:red, 245; green, 166; blue, 35 }  ,fill opacity=1 ] (360.91,415.84) .. controls (360.91,411.99) and (364.36,408.86) .. (368.63,408.86) .. controls (372.89,408.86) and (376.35,411.99) .. (376.35,415.84) .. controls (376.35,419.7) and (372.89,422.82) .. (368.63,422.82) .. controls (364.36,422.82) and (360.91,419.7) .. (360.91,415.84) -- cycle ;
	\draw  [color={rgb, 255:red, 245; green, 166; blue, 35 }  ,draw opacity=1 ][fill={rgb, 255:red, 245; green, 166; blue, 35 }  ,fill opacity=1 ] (360.64,436.46) .. controls (360.64,432.61) and (364.1,429.48) .. (368.37,429.48) .. controls (372.63,429.48) and (376.09,432.61) .. (376.09,436.46) .. controls (376.09,440.32) and (372.63,443.44) .. (368.37,443.44) .. controls (364.1,443.44) and (360.64,440.32) .. (360.64,436.46) -- cycle ;
	\draw  [color={rgb, 255:red, 245; green, 166; blue, 35 }  ,draw opacity=1 ][fill={rgb, 255:red, 245; green, 166; blue, 35 }  ,fill opacity=1 ] (360.91,457.79) .. controls (360.91,453.94) and (364.36,450.81) .. (368.63,450.81) .. controls (372.89,450.81) and (376.35,453.94) .. (376.35,457.79) .. controls (376.35,461.65) and (372.89,464.77) .. (368.63,464.77) .. controls (364.36,464.77) and (360.91,461.65) .. (360.91,457.79) -- cycle ;
	\draw  [color={rgb, 255:red, 245; green, 166; blue, 35 }  ,draw opacity=1 ][fill={rgb, 255:red, 245; green, 166; blue, 35 }  ,fill opacity=1 ] (392.1,403.99) .. controls (392.1,400.14) and (395.55,397.01) .. (399.82,397.01) .. controls (404.08,397.01) and (407.54,400.14) .. (407.54,403.99) .. controls (407.54,407.85) and (404.08,410.97) .. (399.82,410.97) .. controls (395.55,410.97) and (392.1,407.85) .. (392.1,403.99) -- cycle ;
	\draw  [color={rgb, 255:red, 245; green, 166; blue, 35 }  ,draw opacity=1 ][fill={rgb, 255:red, 245; green, 166; blue, 35 }  ,fill opacity=1 ] (392.1,425.32) .. controls (392.1,421.47) and (395.55,418.34) .. (399.82,418.34) .. controls (404.08,418.34) and (407.54,421.47) .. (407.54,425.32) .. controls (407.54,429.18) and (404.08,432.3) .. (399.82,432.3) .. controls (395.55,432.3) and (392.1,429.18) .. (392.1,425.32) -- cycle ;
	\draw  [color={rgb, 255:red, 245; green, 166; blue, 35 }  ,draw opacity=1 ][fill={rgb, 255:red, 245; green, 166; blue, 35 }  ,fill opacity=1 ] (392.36,446.89) .. controls (392.36,443.03) and (395.82,439.91) .. (400.08,439.91) .. controls (404.35,439.91) and (407.8,443.03) .. (407.8,446.89) .. controls (407.8,450.75) and (404.35,453.87) .. (400.08,453.87) .. controls (395.82,453.87) and (392.36,450.75) .. (392.36,446.89) -- cycle ;
	\draw  [color={rgb, 255:red, 245; green, 166; blue, 35 }  ,draw opacity=1 ][fill={rgb, 255:red, 245; green, 166; blue, 35 }  ,fill opacity=1 ] (392.62,467.98) .. controls (392.62,464.13) and (396.08,461) .. (400.34,461) .. controls (404.61,461) and (408.06,464.13) .. (408.06,467.98) .. controls (408.06,471.84) and (404.61,474.96) .. (400.34,474.96) .. controls (396.08,474.96) and (392.62,471.84) .. (392.62,467.98) -- cycle ;
	\draw  [color={rgb, 255:red, 245; green, 166; blue, 35 }  ,draw opacity=1 ][fill={rgb, 255:red, 245; green, 166; blue, 35 }  ,fill opacity=1 ] (424.07,416.32) .. controls (424.07,412.46) and (427.53,409.33) .. (431.8,409.33) .. controls (436.06,409.33) and (439.52,412.46) .. (439.52,416.32) .. controls (439.52,420.17) and (436.06,423.3) .. (431.8,423.3) .. controls (427.53,423.3) and (424.07,420.17) .. (424.07,416.32) -- cycle ;
	\draw  [color={rgb, 255:red, 245; green, 166; blue, 35 }  ,draw opacity=1 ][fill={rgb, 255:red, 245; green, 166; blue, 35 }  ,fill opacity=1 ] (423.81,437.65) .. controls (423.81,433.79) and (427.27,430.66) .. (431.53,430.66) .. controls (435.8,430.66) and (439.26,433.79) .. (439.26,437.65) .. controls (439.26,441.5) and (435.8,444.63) .. (431.53,444.63) .. controls (427.27,444.63) and (423.81,441.5) .. (423.81,437.65) -- cycle ;
	\draw  [color={rgb, 255:red, 245; green, 166; blue, 35 }  ,draw opacity=1 ][fill={rgb, 255:red, 245; green, 166; blue, 35 }  ,fill opacity=1 ] (424.07,458.98) .. controls (424.07,455.12) and (427.53,451.99) .. (431.8,451.99) .. controls (436.06,451.99) and (439.52,455.12) .. (439.52,458.98) .. controls (439.52,462.83) and (436.06,465.96) .. (431.8,465.96) .. controls (427.53,465.96) and (424.07,462.83) .. (424.07,458.98) -- cycle ;
	\draw  [color={rgb, 255:red, 245; green, 166; blue, 35 }  ,draw opacity=1 ][fill={rgb, 255:red, 245; green, 166; blue, 35 }  ,fill opacity=1 ] (455.26,436.94) .. controls (455.26,433.08) and (458.72,429.95) .. (462.99,429.95) .. controls (467.25,429.95) and (470.71,433.08) .. (470.71,436.94) .. controls (470.71,440.79) and (467.25,443.92) .. (462.99,443.92) .. controls (458.72,443.92) and (455.26,440.79) .. (455.26,436.94) -- cycle ;
	\draw    (376.35,415.84) -- (392.1,403.99) ;
	\draw    (376.35,415.84) -- (392.1,425.32) ;
	\draw    (376.35,415.84) -- (392.36,446.89) ;
	\draw    (376.35,415.84) -- (392.21,467.17) ;
	\draw    (376.09,436.46) -- (392.1,403.99) ;
	\draw    (376.09,436.46) -- (392.1,425.32) ;
	\draw    (376.09,436.46) -- (392.36,446.89) ;
	\draw    (376.09,436.46) -- (392.62,467.98) ;
	\draw    (376.35,457.79) -- (392.1,403.99) ;
	\draw    (376.35,457.79) -- (392.1,425.32) ;
	\draw    (376.35,457.79) -- (392.36,446.89) ;
	\draw    (376.35,457.79) -- (392.62,467.98) ;
	\draw    (407.54,403.99) -- (424.49,416.97) ;
	\draw    (407.54,425.32) -- (424.49,416.97) ;
	\draw    (407.8,446.89) -- (424.15,417.83) ;
	\draw    (408.06,467.98) -- (424.49,416.97) ;
	\draw    (407.54,403.99) -- (423.81,436.94) ;
	\draw    (407.54,425.32) -- (423.81,436.94) ;
	\draw    (407.8,446.89) -- (423.81,436.94) ;
	\draw    (408.06,467.98) -- (423.81,436.94) ;
	\draw    (424.07,458.27) -- (407.54,403.99) ;
	\draw    (407.54,425.32) -- (424.07,458.27) ;
	\draw    (407.8,446.89) -- (424.07,458.27) ;
	\draw    (408.06,467.98) -- (424.07,458.27) ;
	\draw    (439.52,415.61) -- (455.26,436.94) ;
	\draw    (439.26,436.94) -- (455.26,436.94) ;
	\draw    (439.52,458.27) -- (455.26,436.94) ;
	\draw    (340.04,389) -- (495,389) -- (495,482.85) -- (340.04,482.85) -- cycle ;
	\draw    (338.52,445.88) .. controls (264.57,474.74) and (334.73,541.93) .. (372,485.32) ;
	\draw [shift={(373.12,483.56)}, rotate = 121.75] [fill={rgb, 255:red, 0; green, 0; blue, 0 }  ][line width=0.08]  [draw opacity=0] (8.93,-4.29) -- (0,0) -- (8.93,4.29) -- cycle    ;
	\draw   (520,285) -- (633,285) -- (633,317) -- (520,317) -- cycle ;
	\draw    (495,441) -- (578,441) ;
	\draw   (577.5,440.5) -- (577.5,323) ;
	\draw [shift={(577.5,320)}, rotate = 89.29] [fill={rgb, 255:red, 0; green, 0; blue, 0 }  ][line width=0.08]  [draw opacity=0] (10.72,-5.15) -- (0,0) -- (10.72,5.15) -- (7.12,0) -- cycle    ;
	\draw  [dash pattern={on 4.5pt off 4.5pt}]  (577.5,189) -- (577.5,279) ;
	\draw  [dash pattern={on 4.5pt off 4.5pt}]  (577.5,181) -- (497.5,181) ;
	\draw [shift={(494.5,180.5)}, rotate = 0.34] [fill={rgb, 255:red, 0; green, 0; blue, 0 }  ][line width=0.08]  [draw opacity=0] (10.72,-5.15) -- (0,0) -- (10.72,5.15) -- (7.12,0) -- cycle    ;
	\draw   (577.5,146) -- (493,146) ;
	\draw   (577.5,91) -- (577.5,147) ;
	\draw   (577.5,92) -- (147.5,92) ;
	\draw      (147.5,91.5) -- (147.5,297.2) ;
	\draw    (147.4,297.2) -- (167,297.2) ;
	\draw [shift={(170,297.2)}, rotate = 180] [fill={rgb, 255:red, 0; green, 0; blue, 0 }  ][line width=0.08]  [draw opacity=0] (10.72,-5.15) -- (0,0) -- (10.72,5.15) -- (7.12,0) -- cycle    ;
	\draw    (147.5,297.2) -- (147.5,442) ;
	\draw   (147.08,441.91) -- (336.08,440.92) ;
	\draw [shift={(339.08,440.91)}, rotate = 179.7] [fill={rgb, 255:red, 0; green, 0; blue, 0 }  ][line width=0.08]  [draw opacity=0] (10.72,-5.15) -- (0,0) -- (10.72,5.15) -- (7.12,0) -- cycle    ;
	\draw  [color={rgb, 255:red, 255; green, 255; blue, 255 }  ,draw opacity=1 ][fill={rgb, 255:red, 255; green, 255; blue, 255 }  ,fill opacity=1 ] (495,204) -- (641,204) -- (641,252) -- (495,252) -- cycle ;
	
	\draw  [cyan, dash pattern={on 4.5pt off 4.5pt}]  (300,56) -- (658,55) ;
	\draw  [cyan, dash pattern={on 4.5pt off 4.5pt}]  (658,55) -- (659,531) ;
	\draw  [cyan, dash pattern={on 4.5pt off 4.5pt}]  (659,531) -- (298,532) ;
	\draw  [cyan, dash pattern={on 4.5pt off 4.5pt}]  (300,56) -- (300,219) ;
	\draw  [cyan, dash pattern={on 4.5pt off 4.5pt}]  (300,219) -- (334,219) ;
	\draw  [cyan, dash pattern={on 4.5pt off 4.5pt}]  (334,219) -- (333,372) ;
	\draw  [cyan, dash pattern={on 4.5pt off 4.5pt}]  (299,372) -- (298,532) ;
	\draw  [cyan, dash pattern={on 4.5pt off 4.5pt}]  (299,372) -- (333,372) ;
	
	\draw (395,330) node [anchor=north west][inner sep=0.75pt]   [align=left] {\footnotesize Samples $\mathcal{D}_{k}$};
	\draw (439.12,183.09) node [anchor=north west][inner sep=0.75pt]   [align=left] {Actor};
	\draw (441.52,461.98) node [anchor=north west][inner sep=0.75pt]   [align=left] {Critic};
	\draw (359.62,498.47) node [anchor=north west][inner sep=0.75pt]   [align=left] {Update};
	\draw (339,66.4) node [anchor=north west][inner sep=0.75pt]    {$A\in \{0,1,2\}$};
	\draw (535,291) node [anchor=north west][inner sep=0.75pt]   [align=left] {Optimizer};
	\draw (180.8,249.68) node [anchor=north west][inner sep=0.75pt]   [align=left] {\footnotesize ADTF Simulation};
	\draw (480,206) node [anchor=north west][inner sep=0.75pt]   [align=left] {Update $\displaystyle \theta $ with the \\gradient $\nabla L_{t}^{CLIP}(\theta)$ };
	\draw (460.62,503.47) node [anchor=north west][inner sep=0.75pt]   [align=left] {High Level RL Agent};

	\end{tikzpicture}
	
	\caption{Reinforcement learning pipeline with ADTF framework}
	\label{fig:training_framework}
\end{figure}

To the best of our knowledge, no relevant literature has described how to integrate reinforcement learning with the ADTF middleware. 
We propose a framework that enables ADTF simulation to interact with the RL agent. 
As is shown in Fig. \ref{fig:training_framework}, the RL agent communicates with the ADTF simulation through TCP/IP communication. 
The action generated by the RL agent is immediately transmitted to ADTF simulation and the observation, for example, states of ego and target vehicle, is generated by the simulation and sent back to the RL agent.
The well-known RL library Gym \cite{brockman2016openai} is used to enable the RL agent to interact with environment and implement RL algorithms. 
Meanwhile, all other algorithms, such as positioning, trajectory planning, and control, are implemented in C++ and packaged as corresponding ADTF filters.
Details of trajectory planning and control are described in  \cite{folkers2020time} \cite{folkers2022opa} \cite{rick2019autonomous}.

\subsection{PPO-based Hierarchical Planning Framework}

\begin{figure}[!tp]
	\centering
	\def \globalscale {0.800000}
	\begin{tikzpicture}[y=0.7cm, x=0.7cm, yscale=\globalscale,xscale=\globalscale, inner sep=0pt, outer sep=0pt]
		\draw[-{Stealth[length=2.5mm, width=1.5mm]}] (3.0, 10.6) -- (3.0, 9.6) -- (3.0, 8.8);      
		\path[draw=cyan,fill=white] (0.9-0.5-0.6, 12.2) rectangle (5.2 + 0.5 + 0.6, 10.6)  node[pos=.5] {High Level RL Agent};      
		\draw[-{Stealth[length=2.5mm, width=1.5mm]}] (3.0,5.4) -- (3.0, 4.3);      
		\path [draw=black,fill=white] (0.0-0.45-1, 8.8) rectangle (6.1 + 0.45 + 1, 5.4) node[midway, yshift=0.7cm]   {Hybrid \text{A}$^{\negthinspace*}$  Trajectory Planner}  node[midway, yshift=-0.2cm] {\includegraphics[width=3cm]{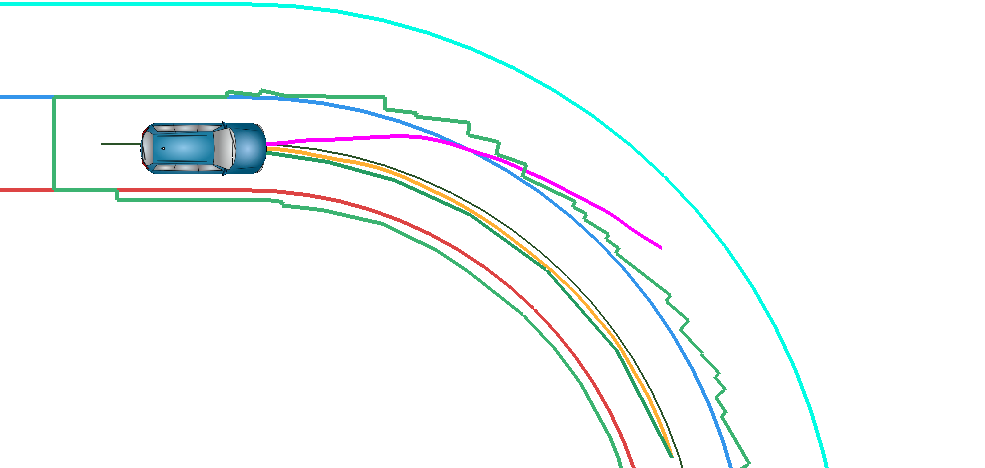}};
		\draw [-{Stealth[length=2.5mm, width=1.5mm]}] (3.0, 2.7) -- (3.0, 1.6);   		
		\path[draw=black,fill=white] (0.9,	4.3) rectangle (5.2, 2.7) node[pos=.5] {MPC}; 
		\draw[-{Stealth[length=2.5mm, width=1.5mm]}] (6.0, 0.8)	-- (10.3, 0.8) -- (10.3, 5+0.1);      
		\path[draw=black,fill=white] (0.1-0.2-0.8, 1.6) rectangle (6.0+0.2+0.8, 0.0) node[pos=.5] {Simulation Env/Real World};
		\draw[-{Stealth[length=2.5mm, width=1.5mm]}] (10.3, 6.6) -- (10.3, 11.4+0.1) -- (5.3+1, 11.4+0.1); 
		\path[draw=black,fill=white] (8.7-0.6,6.6) rectangle (11.9+0.5, 5.1) node[pos=.5] {LTL Reward};
		\node[anchor=east] at (2.9+3.8, 9.75) {$A\in\left\{0,1,2\right\}$};
	\end{tikzpicture}
	\caption{Hierarchical planning framework combining reinforcement learning with trajectory planning and control}
	\label{fig:algorithm_pipeline}
\end{figure}
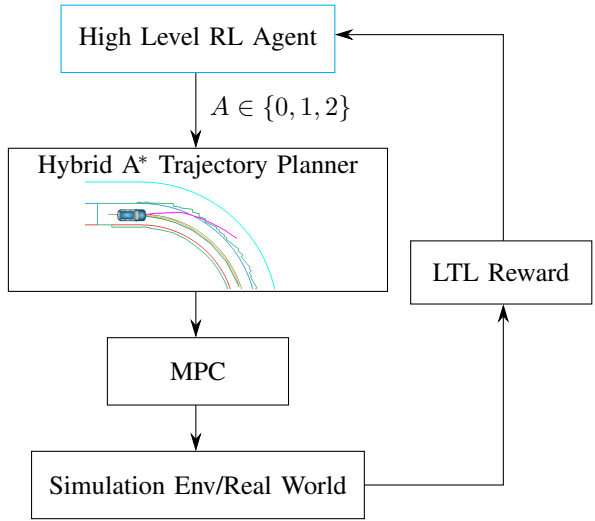

\begin{figure}[t]
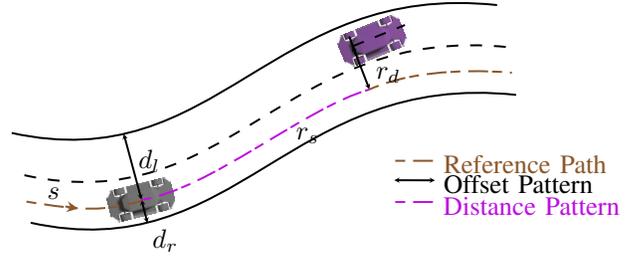


\tikzset{every picture/.style={line width=0.75pt}} 
\hspace*{0.05\linewidth}
\begin{tikzpicture}[x=0.4pt,y=0.4pt,yscale=-1,xscale=1]

\draw (213.83,239.62) node [rotate=-75.78] {\includegraphics[width=11.84pt,height=24.533pt]{figures/car.jpg}};
\draw (432.46,89.08) node [rotate=-67.83] {\includegraphics[width=12.2pt,height=24.533pt]{figures/moving_object_car.jpg}};
\draw  [dash pattern={on 4.5pt off 4.5pt}]  (104.5,216) .. controls (300.5,259) and (348.5,72) .. (563.5,95) ;
\draw    (110.5,261) .. controls (306.5,304) and (354.5,117) .. (569.5,140) ;
\draw    (92.5,176) .. controls (288.5,219) and (336.5,32) .. (551.5,55) ;
\draw [color={rgb, 255:red, 139; green, 87; blue, 42 }  ,draw opacity=1 ]   (143.33,245.6) -- (152.37,247.11) ;
\draw [shift={(155.33,247.6)}, rotate = 189.46] [fill={rgb, 255:red, 139; green, 87; blue, 42 }  ,fill opacity=1 ][line width=0.08]  [draw opacity=0] (10.72,-5.15) -- (0,0) -- (10.72,5.15) -- (7.12,0) -- cycle    ;
\draw    (411.6,84.67) -- (418.4,102.67) ;
\draw  [dash pattern={on 4.5pt off 4.5pt}]  (414.97,93.93) -- (464.4,73.87) ;
\draw [color={rgb, 255:red, 0; green, 0; blue, 0 }  ,draw opacity=1 ]   (199.25,179.75) -- (214.49,236.36) ;
\draw [shift={(215.27,239.25)}, rotate = 254.93] [fill={rgb, 255:red, 0; green, 0; blue, 0 }  ,fill opacity=1 ][line width=0.08]  [draw opacity=0] (5.36,-2.57) -- (0,0) -- (5.36,2.57) -- (3.56,0) -- cycle    ;
\draw [shift={(198.47,176.85)}, rotate = 74.93] [fill={rgb, 255:red, 0; green, 0; blue, 0 }  ,fill opacity=1 ][line width=0.08]  [draw opacity=0] (5.36,-2.57) -- (0,0) -- (5.36,2.57) -- (3.56,0) -- cycle    ;
\draw [color={rgb, 255:red, 139; green, 87; blue, 42 }  ,draw opacity=1 ] [dash pattern={on 3.75pt off 3pt on 7.5pt off 1.5pt}]  (455.22,202.37) -- (491.62,202.37) ;
\draw [color={rgb, 255:red, 0; green, 0; blue, 0 }  ,draw opacity=1 ]   (216.01,242.16) -- (220.13,258.35) ;
\draw [shift={(220.87,261.25)}, rotate = 255.72] [fill={rgb, 255:red, 0; green, 0; blue, 0 }  ,fill opacity=1 ][line width=0.08]  [draw opacity=0] (5.36,-2.57) -- (0,0) -- (5.36,2.57) -- (3.56,0) -- cycle    ;
\draw [shift={(215.27,239.25)}, rotate = 75.72] [fill={rgb, 255:red, 0; green, 0; blue, 0 }  ,fill opacity=1 ][line width=0.08]  [draw opacity=0] (5.36,-2.57) -- (0,0) -- (5.36,2.57) -- (3.56,0) -- cycle    ;
\draw [color={rgb, 255:red, 0; green, 0; blue, 0 }  ,draw opacity=1 ]   (416.05,96.73) -- (429.78,132.06) ;
\draw [shift={(430.87,134.85)}, rotate = 248.77] [fill={rgb, 255:red, 0; green, 0; blue, 0 }  ,fill opacity=1 ][line width=0.08]  [draw opacity=0] (5.36,-2.57) -- (0,0) -- (5.36,2.57) -- (3.56,0) -- cycle    ;
\draw [shift={(414.97,93.93)}, rotate = 68.77] [fill={rgb, 255:red, 0; green, 0; blue, 0 }  ,fill opacity=1 ][line width=0.08]  [draw opacity=0] (5.36,-2.57) -- (0,0) -- (5.36,2.57) -- (3.56,0) -- cycle    ;
\draw [color={rgb, 255:red, 189; green, 16; blue, 224 }  ,draw opacity=1 ] [dash pattern={on 3.75pt off 3pt on 7.5pt off 1.5pt}]  (215.27,239.25) .. controls (279.67,229.65) and (369.67,152.05) .. (430.87,134.85) ;
\draw [color={rgb, 255:red, 139; green, 87; blue, 42 }  ,draw opacity=1 ] [dash pattern={on 3.75pt off 3pt on 7.5pt off 1.5pt}]  (430.87,134.85) .. controls (477.67,113.25) and (546.87,117.25) .. (565.5,119) ;
\draw [color={rgb, 255:red, 139; green, 87; blue, 42 }  ,draw opacity=1 ] [dash pattern={on 3.75pt off 3pt on 7.5pt off 1.5pt}]  (106.5,241) .. controls (160.07,249.85) and (180.47,250.25) .. (215.27,240.25) ;
\draw [color={rgb, 255:red, 0; green, 0; blue, 0 }  ,draw opacity=1 ]   (457.17,220.74) -- (489.87,220.85) ;
\draw [shift={(492.87,220.85)}, rotate = 180.18] [fill={rgb, 255:red, 0; green, 0; blue, 0 }  ,fill opacity=1 ][line width=0.08]  [draw opacity=0] (5.36,-2.57) -- (0,0) -- (5.36,2.57) -- (3.56,0) -- cycle    ;
\draw [shift={(454.17,220.73)}, rotate = 0.18] [fill={rgb, 255:red, 0; green, 0; blue, 0 }  ,fill opacity=1 ][line width=0.08]  [draw opacity=0] (5.36,-2.57) -- (0,0) -- (5.36,2.57) -- (3.56,0) -- cycle    ;
\draw [color={rgb, 255:red, 189; green, 16; blue, 224 }  ,draw opacity=1 ] [dash pattern={on 3.75pt off 3pt on 7.5pt off 1.5pt}]  (455.22,241.37) -- (491.62,241.37) ;

\draw (498.2,194.4) node [anchor=north west][inner sep=0.75pt]  [color={rgb, 255:red, 139; green, 87; blue, 42 }  ,opacity=1 ] [align=left] {Reference Path};
\draw (209.53,189.52) node [anchor=north west][inner sep=0.75pt]    {$d_{l}$};
\draw (123.53,225.12) node [anchor=north west][inner sep=0.75pt]    {$s$};
\draw (222.87,264.65) node [anchor=north west][inner sep=0.75pt]    {$d_{r}$};
\draw (434.33,111.12) node [anchor=north west][inner sep=0.75pt]    {$r_{d}$};
\draw (358.73,168.72) node [anchor=north west][inner sep=0.75pt]    {$r_{s}$};
\draw (498.2,213.4) node [anchor=north west][inner sep=0.75pt]  [color={rgb, 255:red, 0; green, 0; blue, 0 }  ,opacity=1 ] [align=left] {Offset Pattern};
\draw (498,233.6) node [anchor=north west][inner sep=0.75pt]  [color={rgb, 255:red, 189; green, 16; blue, 224 }  ,opacity=1 ] [align=left] {Distance Pattern};
\end{tikzpicture}
	\caption{Observation space in Frenet Coordinates \cite{werling2010optimal}}
	\label{fig:frenet_coordinate}
\end{figure}
\begin{table}
	\medskip
	\begin{center}
		\caption{Defined Action Space}
		\label{tab:action_space}
		\begin{tabular}{c c}
			\hline
			\textbf{Action} & \textbf{Description}\\
			\hline	
			$0$ & Stay in current lane\\
			$1$ & Change to the left lane\\
			$2$ & Change to the right lane\\
			\hline
			\textbf{State} & \textbf{Description}\\
			\hline	
			$r_s$ & Relative longitudinal distance between ego and target vehicle\\
			$r_d$ & Relative lateral distance between ego and target vehicle\\
			$d_l$ & Distance to the left lane boundary of ego vehicle\\
			$d_r$ & Distance to the right lane boundary of ego vehicle\\
			$v$   & Ego vehicle velocity \\
			\hline
		\end{tabular}
	\end{center}
\end{table}
As presented in Fig. \ref{fig:algorithm_pipeline}, a hierarchical planning framework is proposed based on the PPO algorithm. 
Tab. \ref{tab:action_space} defines the required action and observation space.
At timestep $t$, the RL agent receives the observation  $o_t \in \mathcal O = \left\{ r_s, r_d, d_l, d_r, v \right\} $ from ADTF simulation.
To simplify the geometric relationship between the ego and target vehicle, states $ \left\{ r_s, r_d, d_l, d_r \right\}$ are calculated in Frenet coordinates system. 
Fig. \ref{fig:frenet_coordinate} depicts the geometric relationship between the vehicle and lane boundary.
After receiving the observation, the high-level action $a_t \in \mathcal A = \left\{ 0, 1, 2 \right\} $ is generated by the RL agent using the PPO algorithm. Later on it will be denoted as $a^{RL}_t$ to distinguish it from the low-level action  $a^{\text{A}^{\negthinspace*}}_t$.
The optimization objective of PPO  $ L_t^{CLIP+VF+S} \left(\theta \right) $ is given as:
\begin{align}
\begin{split}
L_t^{CLIP+VF+S}  \left(\theta \right) = \hat{\mathbb E}_t & \left[  L_t^{CLIP} \left(\theta \right) - c_1  L_t^{VF} \left(\theta \right) \right. 
\\ & \left.  + c_2  S\left[\pi_\theta \right] \left(s_t \right)  \right],
\label{eq:ppo_objective_function}
\end{split}
\end{align}
where $L_t^{CLIP}$ is the clipped objective (see Eq. \ref{eq:clip_objective_function}), $L_t^{VF}$ is a squared-error loss (see Eq. \ref{eq:error_loss_objective_function}) and $S$ is an entropy bonus. Here $c_1, c_2$ are coefficients and $\theta$ represents the trainable weights of the actor-critic network. The clipped objective and the squared error loss are defined as follows:
\begin{align}
&L_t^{CLIP} \left(\theta \right) =  \min \left( r_t(\theta) \hat{A_t}, \text{clip} \left( r_t(\theta), 1-\epsilon, 1+ \epsilon \right) \hat{A_t} \right) 
\label{eq:clip_objective_function},
\\& L_t^{VF} \left(\theta \right) =  \left( V_\theta (s_t) - V_t^{\text{targ}} \right)^2
\label{eq:error_loss_objective_function},
\end{align}
where $r_t(\theta) = \frac{\pi_\theta\left(a_t| s_t\right)}{\pi_{\theta_{\text{old}}} \left(a_t| s_t\right)}$ denotes the probability ratio. As stated in Eq. \ref{eq:policy_objective_function}, the policy $\pi_\theta$ is generated from the actor-critic network in the current iteration and $\pi_{\theta_{\text{old}}}$ is the old policy before the update. $\hat{A_t}$ is an estimator of the advantage function at timestep $t$, which can be computed by a general advantage estimator \cite{schulman2015high}. The explanation of the function $ \text{clip} \left( r_t(\theta), 1-\epsilon, 1+ \epsilon \right) \hat{A_t}$  can be found in the original paper of PPO \cite{schulman2017proximal}. As for the second loss term $L_t^{VF}$, $V_\theta (s_t)$ is the estimated value function and $V_t^{\text{targ}}$ is the target value function from the critic network.

\subsection{Time Period Reinforcement Learning}
\label{subsec:Time Period Reinforcement Learning}
        
\begin{figure*}[tp]
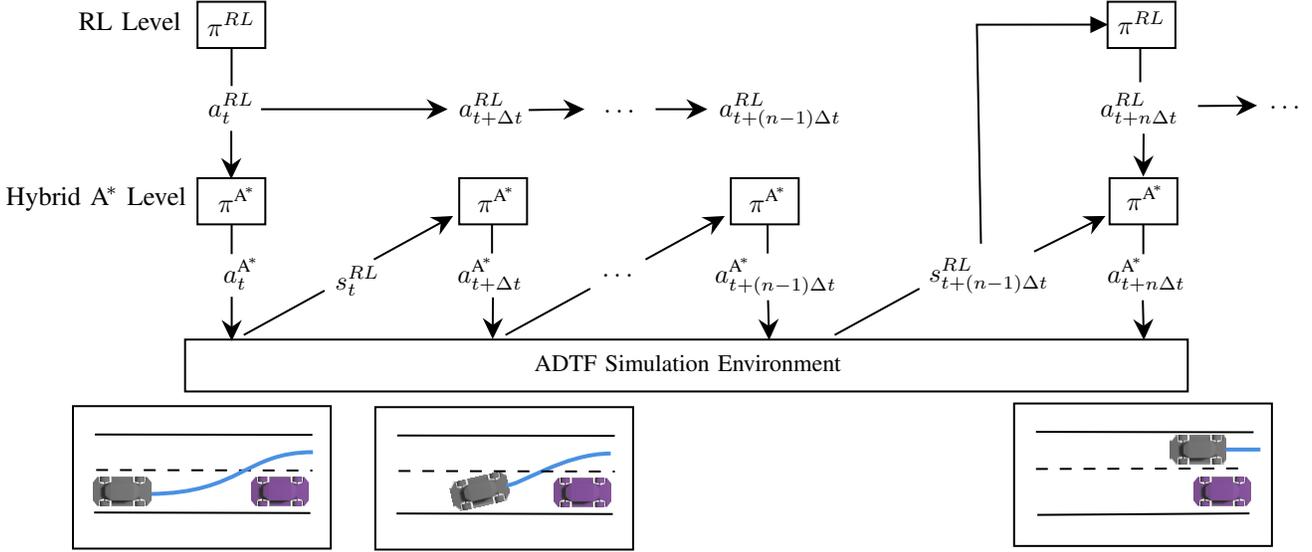


\tikzset{every picture/.style={line width=0.75pt}} 

\begin{tikzpicture}[x=0.75pt,y=0.75pt,yscale=-1,xscale=1]

\draw   (141.79,7) -- (176,7) -- (176,30.34) -- (141.79,30.34) -- cycle ;
\draw    (158.82,72.15) -- (158.82,93.48) ;
\draw [shift={(158.82,96.48)}, rotate = 270] [fill={rgb, 255:red, 0; green, 0; blue, 0 }  ][line width=0.08]  [draw opacity=0] (10.72,-5.15) -- (0,0) -- (10.72,5.15) -- (7.12,0) -- cycle    ;
\draw    (158.99,30.34) -- (158.99,49.52) ;
\draw    (159.04,155.9) -- (159.04,175.24) ;
\draw [shift={(159.04,178.24)}, rotate = 270] [fill={rgb, 255:red, 0; green, 0; blue, 0 }  ][line width=0.08]  [draw opacity=0] (10.72,-5.15) -- (0,0) -- (10.72,5.15) -- (7.12,0) -- cycle    ;
\draw    (159.02,119.17) -- (159.02,134.36) ;
\draw   (135.52,177.49) -- (641.22,177.49) -- (641.22,203.69) -- (135.52,203.69) -- cycle ;
\draw    (165.14,175.18) -- (205.89,152.78) ;
\draw    (230.75,137.09) -- (268.1,116.54) ;
\draw [shift={(270.73,115.1)}, rotate = 151.18] [fill={rgb, 255:red, 0; green, 0; blue, 0 }  ][line width=0.08]  [draw opacity=0] (10.72,-5.15) -- (0,0) -- (10.72,5.15) -- (7.12,0) -- cycle    ;
\draw    (290.79,156.21) -- (290.68,174.27) ;
\draw [shift={(290.66,177.27)}, rotate = 270.35] [fill={rgb, 255:red, 0; green, 0; blue, 0 }  ][line width=0.08]  [draw opacity=0] (10.72,-5.15) -- (0,0) -- (10.72,5.15) -- (7.12,0) -- cycle    ;
\draw    (429.54,157.64) -- (430.01,174.76) ;
\draw [shift={(430.1,177.76)}, rotate = 268.4] [fill={rgb, 255:red, 0; green, 0; blue, 0 }  ][line width=0.08]  [draw opacity=0] (10.72,-5.15) -- (0,0) -- (10.72,5.15) -- (7.12,0) -- cycle    ;
\draw    (536.92,129.09) -- (534.86,18.56) -- (598.03,18.52) ;
\draw [shift={(601.03,18.52)}, rotate = 179.97] [fill={rgb, 255:red, 0; green, 0; blue, 0 }  ][line width=0.08]  [draw opacity=0] (8.93,-4.29) -- (0,0) -- (8.93,4.29) -- cycle    ;
\draw    (618.96,73.61) -- (618.83,92.27) ;
\draw [shift={(618.81,95.27)}, rotate = 270.39] [fill={rgb, 255:red, 0; green, 0; blue, 0 }  ][line width=0.08]  [draw opacity=0] (10.72,-5.15) -- (0,0) -- (10.72,5.15) -- (7.12,0) -- cycle    ;
\draw    (617.26,30.42) -- (617.26,47.61) ;
\draw    (646.38,59.6) -- (673.22,59.4) ;
\draw [shift={(676.22,59.38)}, rotate = 179.58] [fill={rgb, 255:red, 0; green, 0; blue, 0 }  ][line width=0.08]  [draw opacity=0] (10.72,-5.15) -- (0,0) -- (10.72,5.15) -- (7.12,0) -- cycle    ;
\draw    (618.81,119.2) -- (618.81,134.38) ;
\draw   (78.97,211.1) -- (209,211.1) -- (209,283) -- (78.97,283) -- cycle ;
\draw    (90,225.41) -- (199.75,225.6) ;
\draw    (89.67,264.78) -- (198.73,265.06) ;
\draw  [dash pattern={on 4.5pt off 4.5pt}]  (90.69,243.32) -- (199.75,243.6) ;
\draw [color={rgb, 255:red, 74; green, 144; blue, 226 }  ,draw opacity=1 ][line width=1.5]    (117.52,255.43) .. controls (164.4,254.96) and (165.56,234.33) .. (200.03,234.33) ;
\draw    (173.85,61.23) -- (264.98,61.23) ;
\draw [shift={(267.98,61.23)}, rotate = 180] [fill={rgb, 255:red, 0; green, 0; blue, 0 }  ][line width=0.08]  [draw opacity=0] (10.72,-5.15) -- (0,0) -- (10.72,5.15) -- (7.12,0) -- cycle    ;
\draw    (308.8,61.64) -- (333.99,61.27) ;
\draw [shift={(336.98,61.23)}, rotate = 179.17] [fill={rgb, 255:red, 0; green, 0; blue, 0 }  ][line width=0.08]  [draw opacity=0] (10.72,-5.15) -- (0,0) -- (10.72,5.15) -- (7.12,0) -- cycle    ;
\draw    (372.83,61.22) -- (395.98,61.23) ;
\draw [shift={(398.98,61.23)}, rotate = 180.02] [fill={rgb, 255:red, 0; green, 0; blue, 0 }  ][line width=0.08]  [draw opacity=0] (10.72,-5.15) -- (0,0) -- (10.72,5.15) -- (7.12,0) -- cycle    ;
\draw   (141.79,96) -- (176,96) -- (176,119.34) -- (141.79,119.34) -- cycle ;
\draw    (290.33,119.33) -- (290.35,134.03) ;
\draw   (273.79,96) -- (308,96) -- (308,119.34) -- (273.79,119.34) -- cycle ;
\draw   (410.79,96) -- (445,96) -- (445,119.34) -- (410.79,119.34) -- cycle ;
\draw    (367.75,137.09) -- (405.1,116.54) ;
\draw [shift={(407.73,115.1)}, rotate = 151.18] [fill={rgb, 255:red, 0; green, 0; blue, 0 }  ][line width=0.08]  [draw opacity=0] (10.72,-5.15) -- (0,0) -- (10.72,5.15) -- (7.12,0) -- cycle    ;
\draw    (297.14,175.18) -- (339.89,151.78) ;
\draw    (563.75,133.09) -- (597.05,116.44) ;
\draw [shift={(599.73,115.1)}, rotate = 153.43] [fill={rgb, 255:red, 0; green, 0; blue, 0 }  ][line width=0.08]  [draw opacity=0] (10.72,-5.15) -- (0,0) -- (10.72,5.15) -- (7.12,0) -- cycle    ;
\draw    (463.14,175.18) -- (505.89,151.78) ;
\draw   (601.79,96) -- (636,96) -- (636,119.34) -- (601.79,119.34) -- cycle ;
\draw   (600.79,7) -- (635,7) -- (635,30.34) -- (600.79,30.34) -- cycle ;
\draw    (429.33,119.33) -- (429.35,134.03) ;
\draw    (618.54,157.64) -- (619.01,174.76) ;
\draw [shift={(619.1,177.76)}, rotate = 268.4] [fill={rgb, 255:red, 0; green, 0; blue, 0 }  ][line width=0.08]  [draw opacity=0] (10.72,-5.15) -- (0,0) -- (10.72,5.15) -- (7.12,0) -- cycle    ;
\draw (103.79,254.08) node [rotate=-90.95,xslant=-0.01] {\includegraphics[width=10.78pt,height=21.36pt]{figures/car.jpg}};
\draw (283.82,254.21) node [rotate=-77.51,xslant=-0.01] {\includegraphics[width=10.78pt,height=21.36pt]{figures/car.jpg}};
\draw [color={rgb, 255:red, 74; green, 144; blue, 226 }  ,draw opacity=1 ][line width=1.5]    (297.53,251) .. controls (303.53,251.67) and (326.2,235) .. (350.2,235) ;
\draw   (553.63,209.76) -- (683.67,209.76) -- (683.67,281.67) -- (553.63,281.67) -- cycle ;
\draw    (564.66,224.08) -- (674.41,224.27) ;
\draw    (565.35,265.99) -- (672.34,265.24) ;
\draw  [dash pattern={on 4.5pt off 4.5pt}]  (564.66,242.8) -- (670.28,242.52) ;
\draw (646.49,233.21) node [rotate=-89.99,xslant=-0.01] {\includegraphics[width=10.78pt,height=21.36pt]{figures/car.jpg}};
\draw [color={rgb, 255:red, 74; green, 144; blue, 226 }  ,draw opacity=1 ][line width=1.5]    (660.33,233) -- (677.87,233) ;
\draw   (231.47,211.6) -- (361.5,211.6) -- (361.5,283.5) -- (231.47,283.5) -- cycle ;
\draw    (242.5,225.91) -- (352.25,226.1) ;
\draw    (242.17,265.28) -- (351.23,265.56) ;
\draw  [dash pattern={on 4.5pt off 4.5pt}]  (243.19,243.82) -- (352.25,244.1) ;
\draw (183.59,254.33) node [rotate=-89.31,xslant=0.01] {\includegraphics[width=10.78pt,height=21.35pt]{figures/moving_object_car.jpg}};
\draw (336.09,254.83) node [rotate=-89.31,xslant=0.01] {\includegraphics[width=10.78pt,height=21.35pt]{figures/moving_object_car.jpg}};
\draw (658.58,254.05) node [rotate=-89.31,xslant=0.01] {\includegraphics[width=10.78pt,height=21.35pt]{figures/moving_object_car.jpg}};

\draw (145.16,10.79) node [anchor=north west][inner sep=0.75pt]   [align=left] {$\displaystyle \pi^{RL} \ $};
\draw (80.41,10.58) node [anchor=north west][inner sep=0.75pt]   [align=left] {RL Level};
\draw (145.67,51.67) node [anchor=north west][inner sep=0.75pt]   [align=left] {$\displaystyle a_{t}^{RL} \ $};
\draw (150.67,98.81) node [anchor=north west][inner sep=0.75pt]   [align=left] {$\displaystyle \pi^{\text{A}^{\negthinspace*}} \ $};
\draw (43.84,98.81) node [anchor=north west][inner sep=0.75pt]   [align=left] {Hybrid \text{A}$^{\negthinspace*}$  Level};
\draw (310.31,183.91) node [anchor=north west][inner sep=0.75pt]  [font=\small] [align=left] {ADTF Simulation Environment};
\draw (209.76,138.79) node [anchor=north west][inner sep=0.75pt]  [font=\normalsize] [align=left] {$\displaystyle s_{t}^{RL} \ $};
\draw (281.47,98.81) node [anchor=north west][inner sep=0.75pt]   [align=left] {$\displaystyle \pi^{\text{A}^{\negthinspace*}} \ $};
\draw (344.39,142.91) node [anchor=north west][inner sep=0.75pt]    {$\dotsc $};
\draw (509.13,134.79) node [anchor=north west][inner sep=0.75pt]  [font=\normalsize] [align=left] {$\displaystyle s_{t+( n-1) \Delta t}^{RL} \ $};
\draw (604.28,10.15) node [anchor=north west][inner sep=0.75pt]   [align=left] {$\displaystyle \pi^{RL} \ $};
\draw (680.9,58.21) node [anchor=north west][inner sep=0.75pt]    {$\dotsc $};
\draw (153.36,134.79) node [anchor=north west][inner sep=0.75pt]   [align=left] {$\displaystyle a_{t}^{\text{A}^{\negthinspace*}} \ $};
\draw (271.45,134.79) node [anchor=north west][inner sep=0.75pt]   [align=left] {$\displaystyle a_{t+\Delta t}^{\text{A}^{\negthinspace*}} \ $};
\draw (401.32,134.79) node [anchor=north west][inner sep=0.75pt]   [align=left] {$\displaystyle a_{t+( n-1) \Delta t}^{\text{A}^{\negthinspace*}} \ $};
\draw (599.2,134.79) node [anchor=north west][inner sep=0.75pt]   [align=left] {$\displaystyle a_{t+n\Delta t}^{\text{A}^{\negthinspace*}} \ $};
\draw (595.66,51.67) node [anchor=north west][inner sep=0.75pt]   [align=left] {$\displaystyle a_{t+n\Delta t}^{RL} \ $};
\draw (272.16,51.67) node [anchor=north west][inner sep=0.75pt]   [align=left] {$\displaystyle a_{t+\Delta t}^{RL} \ $};
\draw (345.25,59.9) node [anchor=north west][inner sep=0.75pt]    {$\dotsc $};
\draw (403.5, 51.67) node [anchor=north west][inner sep=0.75pt]   [align=left] {$\displaystyle a_{t+( n-1) \Delta t}^{RL} \ $};
\draw (418.47,99.2) node [anchor=north west][inner sep=0.75pt]   [align=left] {$\displaystyle \pi ^{\text{A}^{\negthinspace*}} \ $};
\draw (608.81,98.2) node [anchor=north west][inner sep=0.75pt]   [align=left] {$\displaystyle \pi ^{\text{A}^{\negthinspace*}} \ $};

\end{tikzpicture}

\caption{Time Period Reinforcement Learning with the hybrid \text{A}$^{\negthinspace*}$  path planner }
\label{fig:time_period_rl}
\end{figure*}

The conventional reinforcement learning methods require the RL agent to communicate with the environment in every single step.
For example, for continuous control tasks such as balancing a pendulum or cart-pole, it is reasonable to generate the control signal in each control step. 
However, this poses a problem when it comes to the high-level behavior planning task. 
For instance, the hybrid \text{A}$^{\negthinspace*}$ path planner generates a trajectory to execute that takes more than 100 simulation steps. 
During these simulation steps, the behavior planning command should remain consistent. However, conventional RL methods cannot achieve this goal. 
In order to overcome this problem,  we propose the TPRL method to sample the action in a certain time period and remain consistent in this time period. 
Inspired by \cite{li2021safe} and \cite{gurtler2021hierarchical}, we also divide the policy into two levels. 
One is generated from RL Agent with PPO algorithm and the other is generated from the hybrid \text{A}$^{\negthinspace*}$ path planner.
Although the output of hybrid \text{A}$^{\negthinspace*}$ path planner is not like the RL policy, we consider it as a policy in this hierarchical framework. 
As is shown in Fig. \ref{fig:time_period_rl}, the environment state is partially observed by the RL agent, and the reward and observation are saved in the set of samples $\mathcal{D}_k$ every $N$ time steps. Here we call $N$ as action sampling interval. In this hierarchical framework, the action $a_t^{RL} $ from the RL agent only gives an intermediate behavior planning action. The hybrid \text{A}$^{\negthinspace*}$ trajectory planner actually generates the action $a_t^{\text{A}^{\negthinspace*}}$ and interacts with the environment. 
As for the RL Level, the reward $r^{RL}\left(s^{RL}_{t: t + (N-1)\triangle t}, a^{RL}_{t:  t + (N-1)\triangle t}\right)$ of the TPRL algorithm should be accumulated from time $ t $ until time $t+ (N-1) \triangle t $. The accumulative reward of each high-level decision step is summarized as:

\begin{align}
	r^{RL}\left(s^{RL}_{t: t + (N-1)\triangle t}, a^{RL}_{t:  t + (N-1)\triangle t}\right) = \sum_{n=0}^{N-1}  \gamma^n r\left(s^{\text{A}^{\negthinspace*}}_{t+n\triangle t}, a^{\text{A}^{\negthinspace*}}_{t+n\triangle t}\right) 
	\label{eq:tprl_reward_function}
\end{align}

\begin{algorithm}
	\caption{Training Procedure of TPRL Algorithm }
	\begin{algorithmic}[1]
		\STATE Input: initial policy parameters $\theta_0$, initial value function parameters $\phi_0$
		\FOR {$k = 0,1,2, ...$} 
		\STATE Collect set of samples $\tau \in \mathcal{D}_k = \left\{ \tau_1, \tau_2, ..., \tau_m \right\} $ after running policy $\pi_k = \pi(\theta_k)$ in the environment, where a single sample is \\ $\tau = \left( s^{RL}_t, a^{RL}_t,  r^{RL}\left( s_{t:t+( N-1) \Delta t}^{RL} ,  a_{t:t+( N-1) \Delta t}^{RL}\right)  \right) $
		\STATE Compute rewards-to-go:
		\\$\hat{R_t} = \sum\limits_{t'=t}^{T} r^{RL}\left( s_{t':t'+( N-1) \Delta t}^{RL} ,  a_{t':t'+( N-1) \Delta t}^{RL}\right) $.
		\STATE Compute advantage estimates, $\hat{A_t}$ based on the current value function $V_{ \phi_k } $.
		\STATE Update the policy by maximizing the PPO-Clip objective:
  		\begin{multline*}
  				\theta_{k+1}  =   \arg\max_{\theta}  \frac{1}{|\mathcal{D}_k|  T }   \sum_{\tau \in \mathcal{D}_k}  \sum_{t=0}^{T} \min \left( r_t(\theta) \hat{A_t}, \right.
  		 \\ \left. \text{clip} \left( r_t(\theta), 1-\epsilon, 1+ \epsilon \right) \hat{A_t} \right)  
  	  \end{multline*}
    	\\ via stochastic gradient ascent with Adam.
		\STATE Fit value function by regression on mean-squared error:
		\\ $$\phi_{k+1} = \arg\min_{\phi}   \frac{1}{|\mathcal{D}_k|  T }   \sum_{\tau \in \mathcal{D}_k}  \sum_{t=0}^{T} \left( V_\phi (s_t) - \hat{R_t} \right)^2$$
		\\typically via some gradient descent algorithm.
		\ENDFOR
	\end{algorithmic} 
\end{algorithm}
\subsection{Reward Definition by LTL}
\begin{figure}[t]
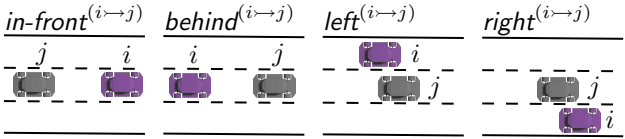

	\centering
	\tikzset{every picture/.style={line width=0.75pt}} 
	
	\begin{tikzpicture}[x=0.3pt,y=0.3pt,yscale=-1,xscale=1]
		
		\draw (91.35,230.65) node [rotate=-89.66] {\includegraphics[width=8.88pt,height=15.512pt]{figures/car.jpg}};
		\draw (201.58,231.18) node [rotate=-89.99] {\includegraphics[width=9.152pt,height=15.6pt]{figures/moving_object_car.jpg}};
		\draw    (53.2,168.6) -- (231,170) ;
		\draw  [dash pattern={on 4.5pt off 4.5pt}]  (53.2,209.6) -- (231,211) ;
		\draw  [dash pattern={on 4.5pt off 4.5pt}]  (53.2,251.6) -- (231,253) ;
		\draw    (53.2,291.6) -- (231,293) ;
		\draw (394.35,232.65) node [rotate=-89.66] {\includegraphics[width=8.88pt,height=15.512pt]{figures/car.jpg}};
		\draw (287.58,232.18) node [rotate=-89.99] {\includegraphics[width=9.152pt,height=15.6pt]{figures/moving_object_car.jpg}};
		\draw    (254.2,169.6) -- (432,171) ;
		\draw  [dash pattern={on 4.5pt off 4.5pt}]  (254.2,210.6) -- (432,212) ;
		\draw  [dash pattern={on 4.5pt off 4.5pt}]  (254.2,252.6) -- (432,254) ;
		\draw    (254.2,292.6) -- (432,294) ;
		\draw (551.35,235.65) node [rotate=-89.66] {\includegraphics[width=8.88pt,height=15.512pt]{figures/car.jpg}};
		\draw (527.58,193.18) node [rotate=-89.99] {\includegraphics[width=9.152pt,height=15.6pt]{figures/moving_object_car.jpg}};
		\draw    (455.2,170.6) -- (633,172) ;
		\draw  [dash pattern={on 4.5pt off 4.5pt}]  (455.2,211.6) -- (633,213) ;
		\draw  [dash pattern={on 4.5pt off 4.5pt}]  (455.2,253.6) -- (633,255) ;
		\draw    (455.2,293.6) -- (633,295) ;
		\draw (752.35,236.65) node [rotate=-89.66] {\includegraphics[width=8.88pt,height=15.512pt]{figures/car.jpg}};
		\draw (779.58,276.18) node [rotate=-89.99] {\includegraphics[width=9.152pt,height=15.6pt]{figures/moving_object_car.jpg}};
		\draw    (656.2,171.6) -- (834,173) ;
		\draw  [dash pattern={on 4.5pt off 4.5pt}]  (656.2,212.6) -- (836,214) ;
		\draw  [dash pattern={on 4.5pt off 4.5pt}]  (656.2,254.6) -- (836,256) ;
		\draw    (656.2,294.6) -- (834,296) ;
		
		\draw (90,174.4) node [anchor=north west][inner sep=0.75pt]    {$j$};
		\draw (199,178.4) node [anchor=north west][inner sep=0.75pt]    {$i$};
		\draw (50,126) node [anchor=north west][inner sep=0.75pt]   [align=left] {${{\fontfamily{cmss}\selectfont{\textit{in-front}}}}^{(i \rightarrowtail j)}$};
		\draw (388,175.4) node [anchor=north west][inner sep=0.75pt]    {$j$};
		\draw (283,178.4) node [anchor=north west][inner sep=0.75pt]    {$i$};
		\draw (251,126) node [anchor=north west][inner sep=0.75pt]   [align=left] {	${{\fontfamily{cmss}\selectfont{\textit{behind}}}}^{(i \rightarrowtail j)}$};
		\draw (585,216.4) node [anchor=north west][inner sep=0.75pt]    {$j$};
		\draw (563,176.4) node [anchor=north west][inner sep=0.75pt]    {$i$};
		\draw (452,126) node [anchor=north west][inner sep=0.75pt]   [align=left] {	${{\fontfamily{cmss}\selectfont{\textit{left}}}}^{(i \rightarrowtail j)}$};
		\draw (786,216.4) node [anchor=north west][inner sep=0.75pt]    {$j$};
		\draw (808,259.4) node [anchor=north west][inner sep=0.75pt]    {$i$};
		\draw (653,126) node [anchor=north west][inner sep=0.75pt]   [align=left] { ${{\fontfamily{cmss}\selectfont{\textit{right}}}}^{(i \rightarrowtail j)}$};

	\end{tikzpicture}
	
	\caption{Vehicle position defined by LTL, where the grey color stands for the ego vehicle and the purple color stands for the target vehicle. }
	\label{fig:ltl_traffic}
\end{figure}

In some situations where RL is applied, the goal is quite clear, so the definition of the reward function is straightforward. However, when applying RL in high-level behavior planning, the reward function can get complicated when the RL agent needs to understand traffic rules. To incorporate the reward function with current traffic rules, LTL is used to formulate the traffic rules logically. We formulate the traffic rules referring to \cite{esterle2020formalizing}.
At first we define $\mathcal{AP} $ as a set of atomic propositions.  The syntax of LTL formula is given by
\begin{align}
\begin{split}
	\varphi :: = &\sigma | \lnot \varphi | \varphi_1 \land \varphi_2 | \varphi_1 \lor \varphi_2 | \varphi_1\Longrightarrow\varphi_2 | 
	\\ &\bigcirc \varphi |   \cup \varphi | \square \varphi |    \lozenge \varphi |,
	\label{eq:ltl_definition_function}
\end{split}
\end{align}
where each atomic proposition $\sigma \in \mathcal{AP}$ is a Boolean statement, $\lnot$,  $\land$, $ \lor $, $\Longrightarrow$  denote the Boolean operators ``not", ``and", ``or", ``implies". Meanwhile, $\bigcirc$,  $\cup$, $ \square $, and $\lozenge$ denote the temporal operators ``next", ``until", ``globally", and ``finally", respectively.
Similarly to \cite{esterle2020formalizing}, we also separate the rules into premise and conclusion,
\begin{align}
	\varphi =  \square(\varphi^p\Longrightarrow\varphi^c)
	\label{eq:premise_conclusion}
\end{align}
where $\varphi^p$ means the current state of the environment or the prerequisite condition. When this condition is fulfilled, the corresponding traffic rule (conclusion) will be checked. $\varphi^c$ denotes the conclusion that defines the legal behavior of the ego agent when the premise is fulfilled. After the LTL format is defined, the formalized rules can be concluded in Tab.
\ref{tab:rules_in_ltl}. Fig. \ref{fig:ltl_traffic} shows some examples of the geometric relationship of LTL labels ${{\fontfamily{cmss}\selectfont{\textit{in-front}}}}^{(i \rightarrowtail j)}$, 	${{\fontfamily{cmss}\selectfont{\textit{behind}}}}^{(i \rightarrowtail j)}$, 	${{\fontfamily{cmss}\selectfont{\textit{left}}}}^{(i \rightarrowtail j)}$ and ${{\fontfamily{cmss}\selectfont{\textit{right}}}}^{(i \rightarrowtail j)}$ between the ego and target vehicle. The  $r_\textnormal{dense}$ denotes the distance between the ego and surrounding vehicles.

\begin{table}[t]
	\begin{center}
		\caption{Atomic propositions and interpretations}
		\label{tab:rules_in_ltl}
		\begin{tabular}{c c}
			\hline
			\noalign{\vskip 0.5mm} 
			$ \sigma \in \mathcal{AP} $ & Interpretation\\
			\noalign{\vskip 0.5mm} 
			\hline	
			\noalign{\vskip 0.5mm} 
			${{\fontfamily{cmss}\selectfont{\textit{dense}}}}^{(i)}$ & $i$ is closer than $r_\textnormal{dense}$ to specific number of vehicles\\
			${{\fontfamily{cmss}\selectfont{\textit{right}}}}^{(i \rightarrowtail j)}$ & $i$ is to the right of $j$\\
			${{\fontfamily{cmss}\selectfont{\textit{left}}}}^{(i \rightarrowtail j)}$ &  $i$ is to the left of $j$\\
			${{\fontfamily{cmss}\selectfont{\textit{in-front}}}}^{(i \rightarrowtail j)}$ &  $i$ is in the front of $j$\\
			${{\fontfamily{cmss}\selectfont{\textit{behind}}}}^{(i \rightarrowtail j)}$ &  $i$ is behind of $j$ \\
			${{\fontfamily{cmss}\selectfont{\textit{sd-front}}}}^{(i)}$ & $i$ has a safe distance to the preceding vehicle \\
			${{\fontfamily{cmss}\selectfont{\textit{sd-rear}}}}^{(i)}$ &  $i$ has a safe distance to the following vehicle\\
			${{\fontfamily{cmss}\selectfont{\textit{lane-change}}}}^{(i)}$ & $i$ is crossing a lane boundary\\
			${{\fontfamily{cmss}\selectfont{\textit{rightmost-lane}}}} ^{(i)}$ & $i$ is in the rightmost lane \\
			\noalign{\vskip 0.5mm} 
			\hline
		\end{tabular}
	\end{center}
\end{table}

\begin{table}
	\begin{center}
		\caption{TRAFFIC RULES FORMULATED BY LTL}
		\label{tab:formalized_rules_in_ltl}
		\begin{tabular}{c c}
			\hline
			\noalign{\vskip 0.5mm} 
			Rule  & Formula $\varphi$  \\
			\noalign{\vskip 0.5mm} 
			\hline	
			\noalign{\vskip 0.5mm} 
			$R_1$: keep in right most lane &  $\square\left(\lnot{{\fontfamily{cmss}\selectfont{\textit{dense}}}} ^{(i)} \Longrightarrow {{\fontfamily{cmss}\selectfont{\textit{rightmost-lane}}}} ^{(i)} \right)$ \\
			$R_2$: safe lane change & 	$\square\left(  {{\fontfamily{cmss}\selectfont{\textit{lane-change}}}}  \Longrightarrow {{\fontfamily{cmss}\selectfont{\textit{sd-front}}}} ^{(i)} \right)$ \\
			\noalign{\vskip 0.5mm} 
			\hline
		\end{tabular}
	\end{center}
\end{table}

Based on these basic labels, the required traffic rules can be formalized in the format of premise and conclusion. In Tab. \ref{tab:formalized_rules_in_ltl}, we summarize two rules that can be used as the reward terms. When the ego agent violates $R_1$ or $R_2$ once, the reward will be $-1$  for the current timestep. If it fulfills the traffic rules, then the reward is to be $0$, indicating that it will not get punished when it obeys the traffic rules. 
Therefore, the reward can be written as
\begin{align}
	\begin{split}
		 r^{RL}\left(s^{RL}_{t}, a^{RL}_{t}\right) = R_1 + R_2
		\label{eq:reward_inltl_definition}
	\end{split}
\end{align}

\section{Experiments and Results}
\label{sec:results}

To demonstrate the performance of the proposed TPRL algorithm, we designed several experiments based on our simulation environment and also validated it in real-world situations. The entire training and testing process is conducted on a Lenovo Thinkpad T14 Laptop with an AMD Ryzen 7 PRO 4750U CPU kernel, accompanied by AMD Radeon Graphics. Each trial requires approximately 24 hours for complete training. The real-world experiments are conducted using two ADAS model cars, and the hardware configuration details can be referenced in \cite{li2023indoor}. A fully-connected multi-layer perceptron network with two hidden layers is employed to represent the policy, comprising 64 units in the first layer and 32 units in the second layer. The implemented code related to the RL aspect is based on OpenAI's Spinning Up \cite{SpinningUp2018}. The entire trajectory planning and control algorithm is implemented in C++ with the ADTF middleware while the RL algorithm is developed in python using the TensorFlow platform \cite{folkers2022opa}.  

\subsection{Experimental Setup}
\label{subsection:Experimental_Setup}

\textit{ 1) Training setups: } To validate the generalization ability of the proposed algorithm, we train our RL agent on a very simple oval map and validate it with a more complex cross road map in ADTF simulation. As is shown in Fig. \ref{fig:oval_map}, the target vehicle is initialized with 0.278 m/s (1 km/h), and after 8 seconds the ego vehicle is initialized with 0.556 m/s (2 km/h). Due to the time gap, the target vehicle is in front of the ego vehicle in the very beginning. Then, the ego vehicle will consistently overtake the target vehicle because the target vehicle is slower than the ego vehicle. The purple trajectory in Fig. \ref{fig:oval_map} is generated when the ego agent receives the lane change decision from our RL agent. Otherwise, the ego vehicle will follow the target vehicle in front with 0.278 m/s. Thanks to our hybrid \text{A}$^{\negthinspace*}$ path planner \cite{folkers2020time} and the emergency stop function, a collision-free trajectory is ensured. The ego vehicle drives 5 circles along the map for each episode. Referring to \cite{SpinningUp2018} and \cite{mysore2021regularizing}, we also implemented a similar RL framework with the gym library. The selection of hyperparameter parameters is also similar to \cite{SpinningUp2018} and we list it in the Tab. \ref{tab:ppo_parameters}. 

\begin{table}[t]
	\begin{center}
		\caption{Hyperparameters used in PPO algorithm}
		\label{tab:ppo_parameters}
		\begin{tabular}{c c}
			\hline
			\noalign{\vskip 0.5mm} 
			Parameters  & Value \\
			\noalign{\vskip 0.5mm} 
			\hline	
			\noalign{\vskip 0.5mm} 
			Policy optimizer learning rate & $1e^{-3}$\\
			Number of gradient descent steps & 80\\
			Value function optimizer learning rate & $3e^{-4}$\\
			PPO clip ratio & 0.2\\
			Discount factor $\lambda$ & 0.99 \\
			Minibatch size & 46 \\
			Steps per episodes & 20000 \\
			Action sampling interval $N$ & 300 \\
			Advantage estimation discount factor &  0.97 \\
		    Neural network nonlinearity & $tanh$\\
			\noalign{\vskip 0.5mm} 
			\hline
		\end{tabular}
	\end{center}
\end{table}
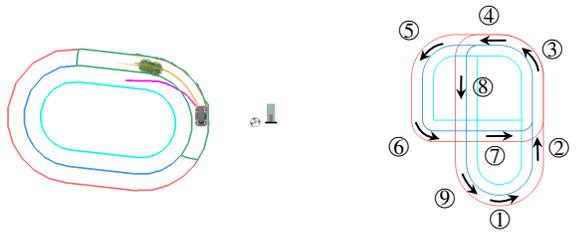
\begin{figure}[t]
	\centering	
	\subfloat[Training settings in oval map, where the vehicle shaded with the two green circles is the target vehicle.]{\centering
		\includegraphics[width=0.5\linewidth]{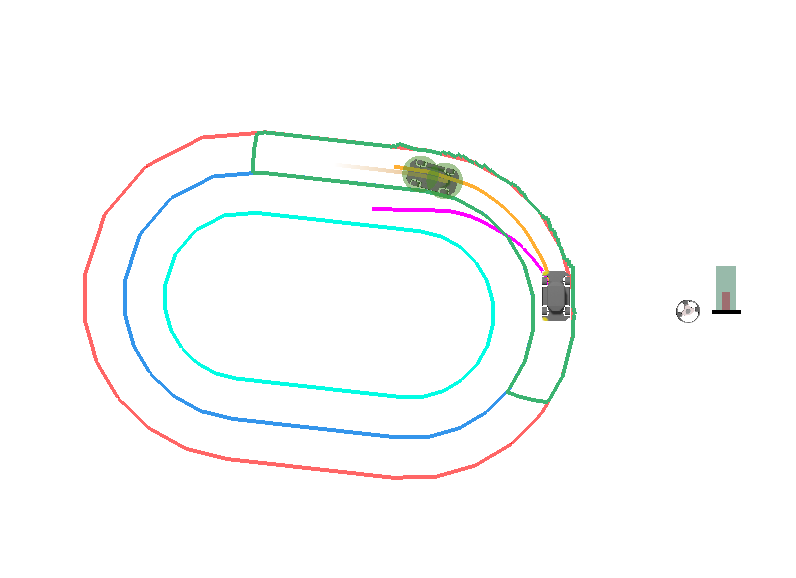}
		\label{fig:oval_map}}
	\qquad
	\subfloat[Test settings in cross road map, where those arrows show the path of ego vehicle. ]{
		\centering

	\tikzset{every picture/.style={line width=0.75pt}} 
	
	\begin{tikzpicture}[x=0.75pt,y=0.75pt,yscale=-1,xscale=1]
	
	\draw (180.21,78.49) node  {\includegraphics[width=66.57pt,height=71.03pt]{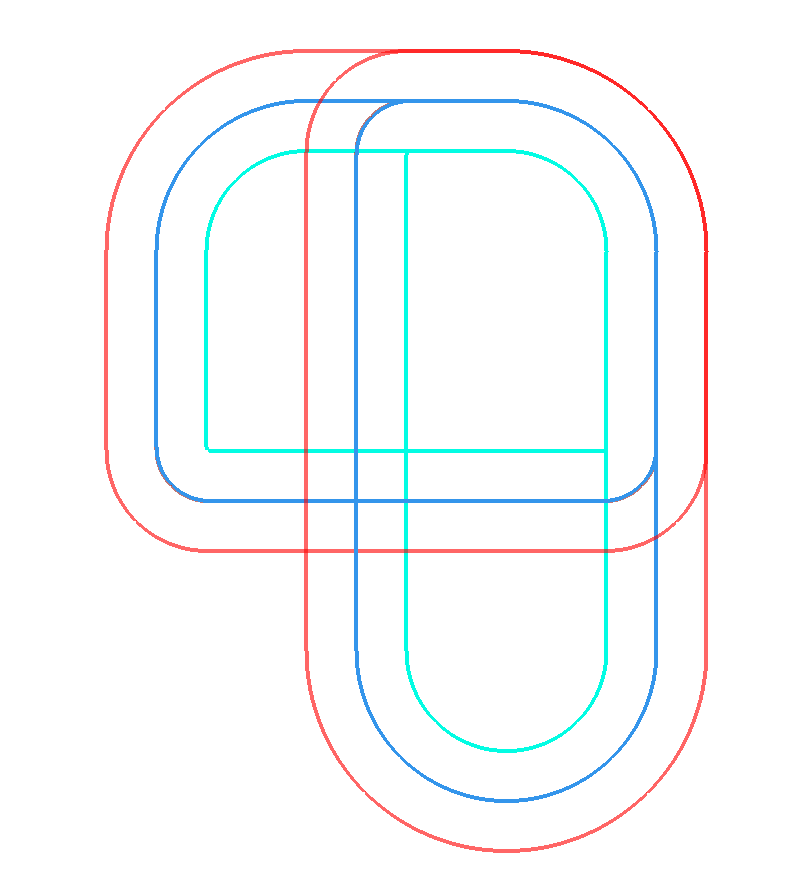}};
	\draw    (187.21,119.76) .. controls (191.68,121.43) and (195.85,120.13) .. (198.95,118.89) ;
	\draw [shift={(201.71,117.76)}, rotate = 159.48] [fill={rgb, 255:red, 0; green, 0; blue, 0 }  ][line width=0.08]  [draw opacity=0] (5.36,-2.57) -- (0,0) -- (5.36,2.57) -- (3.56,0) -- cycle    ;
	\draw    (172.71,57.01) -- (172.6,61.58) -- (172.6,66.01) ;
	\draw [shift={(172.61,69.01)}, rotate = 269.94] [fill={rgb, 255:red, 0; green, 0; blue, 0 }  ][line width=0.08]  [draw opacity=0] (5.36,-2.57) -- (0,0) -- (5.36,2.57) -- (3.56,0) -- cycle    ;
	\draw    (173.21,106.51) .. controls (174.72,110.2) and (176.8,112.96) .. (178.78,115) ;
	\draw [shift={(180.96,117.01)}, rotate = 219.68] [fill={rgb, 255:red, 0; green, 0; blue, 0 }  ][line width=0.08]  [draw opacity=0] (5.36,-2.57) -- (0,0) -- (5.36,2.57) -- (3.56,0) -- cycle    ;
	\draw    (195.33,39.24) -- (190.81,39.27) -- (184.87,39.3) ;
	\draw [shift={(181.87,39.31)}, rotate = 359.72] [fill={rgb, 255:red, 0; green, 0; blue, 0 }  ][line width=0.08]  [draw opacity=0] (5.36,-2.57) -- (0,0) -- (5.36,2.57) -- (3.56,0) -- cycle    ;
	\draw    (185.21,87.57) -- (189.79,87.53) -- (195.83,87.48) ;
	\draw [shift={(198.83,87.45)}, rotate = 179.47] [fill={rgb, 255:red, 0; green, 0; blue, 0 }  ][line width=0.08]  [draw opacity=0] (5.36,-2.57) -- (0,0) -- (5.36,2.57) -- (3.56,0) -- cycle    ;
	\draw    (211.57,54.82) .. controls (211.01,50.08) and (207.94,46.97) .. (205.42,44.79) ;
	\draw [shift={(203.15,42.85)}, rotate = 42.19] [fill={rgb, 255:red, 0; green, 0; blue, 0 }  ][line width=0.08]  [draw opacity=0] (5.36,-2.57) -- (0,0) -- (5.36,2.57) -- (3.56,0) -- cycle    ;
	\draw    (163.59,40.98) .. controls (158.86,41.59) and (155.78,44.69) .. (153.62,47.24) ;
	\draw [shift={(151.71,49.53)}, rotate = 311.6] [fill={rgb, 255:red, 0; green, 0; blue, 0 }  ][line width=0.08]  [draw opacity=0] (5.36,-2.57) -- (0,0) -- (5.36,2.57) -- (3.56,0) -- cycle    ;
	\draw    (149.71,80.51) .. controls (152.1,86.57) and (155.79,87.79) .. (158.81,88.4) ;
	\draw [shift={(161.71,89.01)}, rotate = 197.38] [fill={rgb, 255:red, 0; green, 0; blue, 0 }  ][line width=0.08]  [draw opacity=0] (5.36,-2.57) -- (0,0) -- (5.36,2.57) -- (3.56,0) -- cycle    ;
	\draw    (211.35,100.01) -- (211.27,95.44) -- (211.08,91.01) ;
	\draw [shift={(210.96,88.01)}, rotate = 87.55] [fill={rgb, 255:red, 0; green, 0; blue, 0 }  ][line width=0.08]  [draw opacity=0] (5.36,-2.57) -- (0,0) -- (5.36,2.57) -- (3.56,0) -- cycle    ;
	
	\draw (185.05,124.39) node [anchor=north west][inner sep=0.75pt]  [font=\footnotesize]  {$\textcircled{\raisebox{-0.9pt}{1}} $};
	\draw (214.87,88.05) node [anchor=north west][inner sep=0.75pt]  [font=\footnotesize]  {$\textcircled{\raisebox{-0.9pt}{2}} $};
	\draw (211.53,38.44) node [anchor=north west][inner sep=0.75pt]  [font=\footnotesize]  {$\textcircled{\raisebox{-0.9pt}{3}} $};
	\draw (179.59,21.48) node [anchor=north west][inner sep=0.75pt]  [font=\footnotesize]  {$\textcircled{\raisebox{-0.9pt}{4}} $};
	\draw (139.31,29.13) node [anchor=north west][inner sep=0.75pt]  [font=\footnotesize]  {$\textcircled{\raisebox{-0.9pt}{5}} $};
	\draw (134.1,87.84) node [anchor=north west][inner sep=0.75pt]  [font=\footnotesize]  {$\textcircled{\raisebox{-0.9pt}{6}} $};
	\draw (182.64,92.44) node [anchor=north west][inner sep=0.75pt]  [font=\footnotesize]  {$\textcircled{\raisebox{-0.9pt}{7}} $};
	\draw (176.54,57.35) node [anchor=north west][inner sep=0.75pt]  [font=\footnotesize]  {$\textcircled{\raisebox{-0.9pt}{8}} $};
	\draw (157.36,113.57) node [anchor=north west][inner sep=0.75pt]  [font=\footnotesize]  {$\textcircled{\raisebox{-0.9pt}{9}} $};

	\end{tikzpicture}

		\label{fig:cross_road_map}
	}
	\caption{Maps used for training and test enviroments}
\end{figure}

\textit{ 2) Reference algorithm for comparison: }Although combining a rule-based state machine with the hybrid \text{A}$^{\negthinspace*}$ path planner could achieve better results, when it adapts to our test scenario. It lacks scalability as the scenario becomes more complex. The goal of this paper is to validate whether the reinforcement learning algorithm can be combined with the hybrid \text{A}$^{\negthinspace*}$ path planner and to evaluate its performance. Therefore, another state-of-the-art RL algorithm, Double DQN (DDQN), is implemented and benchmarked as the reference algorithm. Conventionally, the RL Agent will interact with the hybrid \text{A}$^{\negthinspace*}$ path planner in every single simulation step. In our case, to validate the performance of the TPRL algorithm, we provide three time skip variants for benchmarking:
\begin{itemize}
	\item No skip: Every action generated by the RL agent will be used as input for the ADTF Simulation. However, to maintain the consistency of the trajectory, the action will be refused when the hybrid \text{A}$^{\negthinspace*}$ path planner has generated the trajectory. All state-action pairs will be saved in the set of samples $ \mathcal{D}_k$.
	\item Period skip: Similar to what is stated in \cite{wu2017flow}, the no skip method could lead to an excessive number of lane changes,  which can confuse the path planner. Therefore, a large lane change duration is required for the RL agent. The RL agent will skip the action for a certain time period and only transmit new lane-changing commands after this duration. In our case, the time period is configured as 300 simulation time steps because the execution time of a single trajectory generated from hybrid \text{A}$^{\negthinspace*}$  is around this number of simulation time steps. All state-action pairs generated during this time period will be saved in the set of samples $ \mathcal{D}_k$.
	\item TPRL: This is similar to the period skip but only the accumulated reward and observation every 300 time steps will be saved in the set of samples $ \mathcal{D}_k$.
\end{itemize}

\textit{ 3) Test setups and metrics: }After completing the training process, the model is frozen and used for the test scenario. The test is conducted on the cross road map in Fig. \ref{fig:cross_road_map} and evaluated using three metrics:
\begin{itemize}
	\item Traffic rule compliance: This measures the ratio of the simulation timesteps where the ego vehicle follows the traffic rules to the total simulation timesteps.
	\item Emergency brake rate: This indicates how often the ego agent activates the emergency brake during the entire test.
	\item Time costs: This refers to the overall time consumption when the ego agent completes the entire test.
\end{itemize}
We implement the test scenario similarly to the training setups. The target vehicle is initialized with 0.278 m/s, and after 8 seconds, the ego vehicle is initialized with 0.556 m/s. The ego vehicle drives 10 times along the cross road map and is evaluated using the aforementioned metrics.

\begin{figure}[t]
	\centering
	\includegraphics[width=1\linewidth]{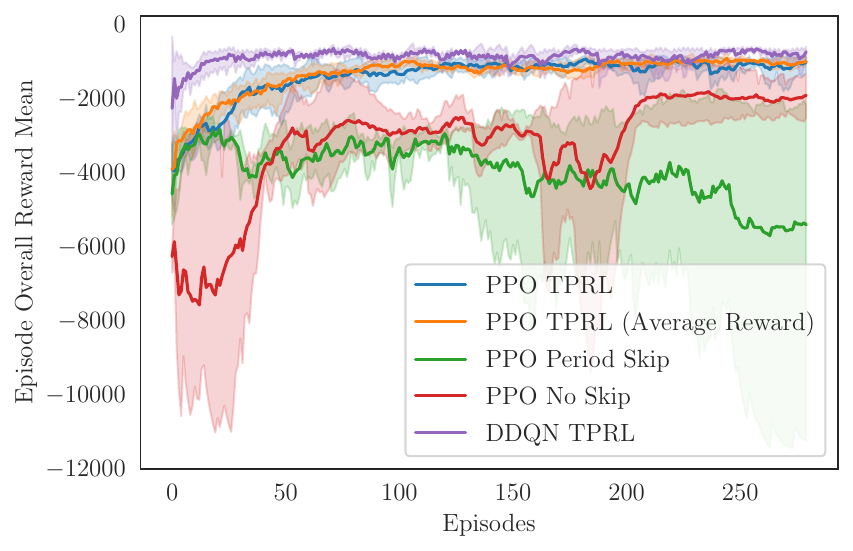}
	\caption{Overall reward mean of PPO and DDQN methods with different settings mentioned in subsection \ref{subsection:Experimental_Setup}}
	\label{fig:overall_reward}
\end{figure}

\begin{figure}[t]
	\centering	
	\subfloat[Reward for the rule $R_1$]{
		\includegraphics[width=0.43\linewidth]{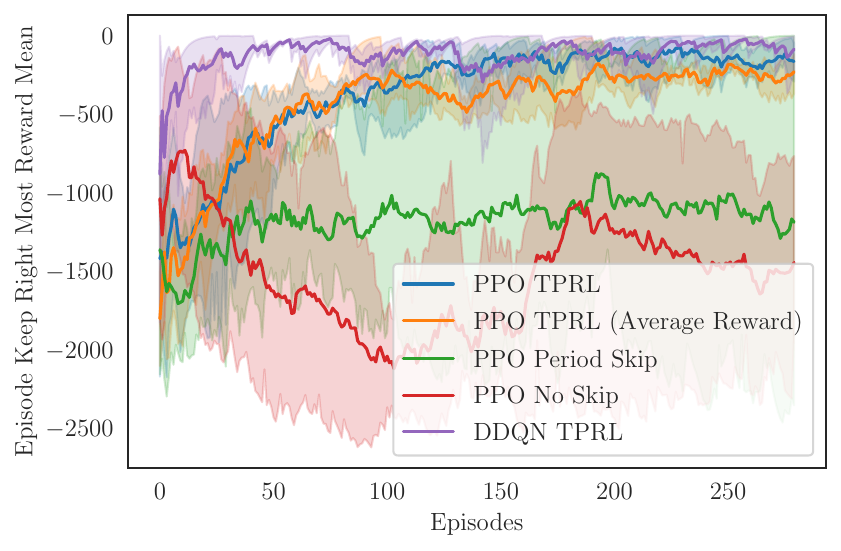}
		\label{fig:keep_right_most_reward}}
	\qquad
	\subfloat[Reward for the rule $R_2$]{
		\includegraphics[width=0.43\linewidth]{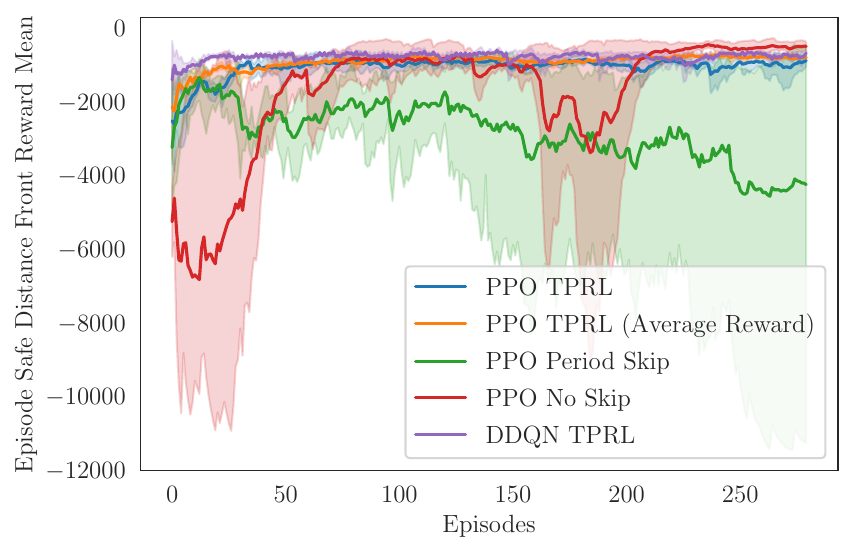}
		\label{fig:sd_front_reward}}
	\caption{Separate reward curve of LTL defined reward}
	\label{fig:r1_and_r2}
\end{figure}

\subsection{Results and Discussions}

\textit{ 1) Training results: } As stated in subsection \ref{subsection:Experimental_Setup}, we train our RL agent with DDQN and PPO based on three time skip variants and each is trained with three random seeds. In Fig  \ref{fig:overall_reward} and Fig. \ref{fig:r1_and_r2}, the colored lines represent the reward mean of five training results where the shaded areas indicating the standard deviation. Fig. \ref{fig:overall_reward} shows the overall reward mean curve $r^{RL}(s^{RL}_{t}, a^{RL}_{t})$ of DDQN and PPO respectively. For PPO methods, agents trained with no skip and period skip exhibit considerable variance compared to the TPRL algorithm, which indicates that the training process is not so stable. Conversely, with the TPRL algorithm, the agent follows traffic rules better and achieves a higher overall reward mean. It's essential to highlight that a large negative reward indicates frequent violations of defined traffic rules by the ego vehicle. Furthermore, Fig. \ref{fig:keep_right_most_reward} and Fig. \ref{fig:sd_front_reward} represent the trained reward mean for each rule. Due to frequent lane change decisions, the no skip method achieves a slightly higher reward for the rule $R_2$. However, the rule $R_1$ is not well obeyed because of the frequent lane change. Therefore, we could conclude that the TPRL algorithm outperforms the other two algorithms in terms of sample efficiency. The reason behind this is that, in TPRL the observation and reward are efficiently sampled from the environment. In contrast, in the period skip method, all state action pairs and rewards are stored in the set of samples, where only the action in every 300 time steps is executed in the environment.

\begin{figure}[t]
	\centering
	\includegraphics[width=1\linewidth]{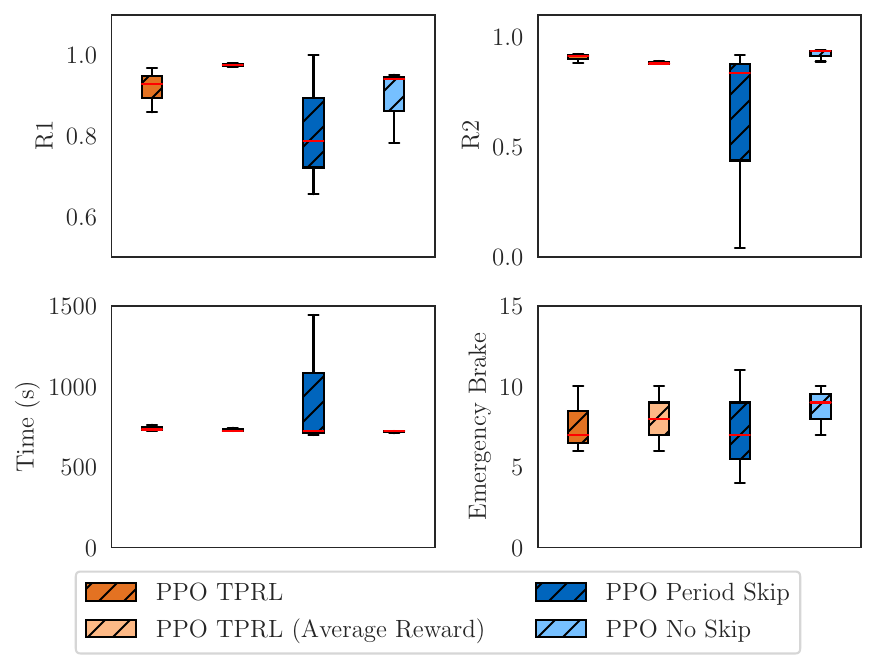}
	\caption{Test results in cross road map based on three metrics}
	\label{fig:test_results}
\end{figure}

\textit{ 2) Test results: } Fig. \ref{fig:test_results} shows test results of the PPO algorithm with three time skip variants. Training results based on DDQN and PPO have confirmed that the TPRL algorithm achieves adequate performance for both algorithms. Therefore, we only compare test results based on PPO algorithm. The RL algorithm is trained with three random seeds, and thus, we test it with three inferences generated by those seeds in the testing scenario. Here the box plot is used to illustrate the median, maximum, and minimum values of three random seeds. The red line in Fig. \ref{fig:test_results} represents the median value of three inferences, while the points, where the horizontal lines terminate, denote the upper and lower values of test results based on three metrics.

From the box plot, we can conclude that both of our agents, based on the TPRL algorithm with accumulated or average reward, achieve a sufficient rule compliance rate when finishing the test scenario. Compared to the algorithm with period skip or no skip, the rule compliance rate is higher and the variance is lower because the lane change action is sufficiently executed. The agent efficiently learns the safe lane change rules because we use not the Cartesian coordinates but the Frenet coordinates. The relative geometric relationship based on the Frenet coordinates leads to the effective transfer of the learned policy. 

For the period skip method, the results vary considerably  with rule $R_2$, because it is not well obeyed in one test. The test results align with the reward curves observed during the training phase. In Fig  \ref{fig:overall_reward} the reward curve of period skip exhibits large variance and meanwhile, in Fig \ref{fig:sd_front_reward} we observe that the period skip method does not consistently adhere to the $R_2$. In one test round, we find that the RL agent almost only follows the target vehicle and makes fewer lane change decisions. Therefore the lowest $R_2$ value of period skip is far from the standard. Throughout the training and test process, our TPRL algorithm demonstrates greater stability than the period skip and achieves a better rule compliance rate. 

Regarding the no skip method, we observe that lane change commands are executed at a very high frequency which is initially rejected by our planner. Consequently, the $R_1$ is not adequately obeyed. The lower bound of $R_1$ reaches 0.78 whereas TPRL consistently exceeds 0.85. The emergency brake caused by TPRL is also less than no skip method. In summary, the performance of the TPRL algorithm exceeds the no skip or period skip methods. With the LTL defined reward, the RL algorithm can achieve reasonable lane change decisions and learn the defined traffic rules. In addition, we find that by averaging the reward, more stable training and test results can be achieved.

\section{CONCLUSIONS AND FUTURE WORK}
\label{sec:conclusion}
We present a hierarchical decision-making algorithm that combines DRL with the Hybrid \text{A}$^{\negthinspace*}$ trajectory planner. Therein, the time period reinforcement learning is proposed in order to sample a single action in a fixed time period. This method maintains consistency in lane change actions, enabling the local trajectory planner to successfully plan and execute trajectories within specific time periods. Furthermore, LTL rules are implemented as reward functions and tested. We find that the agent is able to follow the traffic rules based on LTL and the TPRL algorithm, whereas the simple period skip or no skip methods fail to achieve a sufficient rule compliance rate. In addition, we also implement the algorithm in a real-world scenario and validate its real-time feasibility. Future work will involve utilizing a bird's-eye view as input and scaling up the scenario, testing it in more complex real-world traffic situations.




%

\section{ACKNOWLEDGEMENT}
\label{sec:acknowledgement}

The work of Xibo Li is supported by the Federal Ministry of Economic Affairs and Climate Action on the basis of a decision by the German Bundestag. The work of Shruti Patel is supported by funds of the German Governments Special Purpose Fund held at Landwirtschaftliche Rentenbank and the European Regional Development Fund (ERDF). We thank our colleague Matthias Rick and our former colleague Dr. Andreas Folkers for providing the trajectory planning and control software module.

\bibliographystyle{IEEEtran}
\bibliography{IEEEabrv,references}

\end{document}